\documentclass[letterpaper]{article} % DO NOT CHANGE THIS
\usepackage[submission]{aaai24}  % DO NOT CHANGE THIS
\usepackage{times}  % DO NOT CHANGE THIS
\usepackage{helvet}  % DO NOT CHANGE THIS
\usepackage{courier}  % DO NOT CHANGE THIS
\usepackage[hyphens]{url}  % DO NOT CHANGE THIS
\usepackage{graphicx} % DO NOT CHANGE THIS
\usepackage{natbib}  % DO NOT CHANGE THIS AND DO NOT ADD ANY OPTIONS TO IT
\usepackage{caption} % DO NOT CHANGE THIS AND DO NOT ADD ANY OPTIONS TO IT
\usepackage{algorithm}
\usepackage{algorithmic}
\usepackage{graphicx}
\usepackage{amsmath}
\usepackage{booktabs}
\usepackage{xcolor}
\usepackage{paralist}
\usepackage{subcaption}
\usepackage{multirow}
\usepackage{refcount}

\usepackage{newfloat}
\usepackage{listings}
\DeclareCaptionStyle{ruled}{labelfont=normalfont,labelsep=colon,strut=off} % DO NOT CHANGE THIS
\floatstyle{ruled}
\newfloat{listing}{tb}{lst}{}
\floatname{listing}{Listing}
\title{Breaking the Global North Stereotype:\\ A Global South-centric Benchmark Dataset for Auditing\\ and Mitigating Biases in Facial Recognition Systems}
\author {
    Siddharth Jaiswal\textsuperscript{\rm 1},
    Animesh Ganai\textsuperscript{\rm 1},
    Abhisek Dash\textsuperscript{\rm 2},
    Saptarshi Ghosh\textsuperscript{\rm 1},
    Animesh Mukherjee\textsuperscript{\rm 1}
}
\affiliations {
    \textsuperscript{\rm 1}Indian Institute of Technology Kharagpur, India\\
    \textsuperscript{\rm 2}Max Planck Institute for Software Systems, Germany\\
}
\begin{document}

\maketitle

\begin{abstract}
Facial Recognition Systems (FRSs) are being developed and deployed all around the world at unprecedented rates. Most platforms are designed in a limited set of countries, but deployed in other regions too, without adequate checkpoints for region-specific requirements. This is especially problematic for Global South countries which lack strong legislation to safeguard persons facing disparate performance of these systems. A combination of unavailability of datasets, lack of understanding of how FRSs function and low-resource bias mitigation measures accentuate the problems at hand. In this work, we propose a self-curated face dataset composed of 6,579 unique male and female sports-persons (cricket players) from eight countries around the world. More than 50\% of the dataset is composed of individuals from the Global South countries and is demographically diverse. To aid adversarial audits and robust model training, we curate four adversarial variants of each image in the dataset, leading to more than 40,000 distinct images. We also use this dataset to benchmark five popular facial recognition systems (FRSs), including both commercial and open-source FRSs, for the task of gender prediction (and country prediction for one of the open-source models as an example of red-teaming). Experiments on industrial FRSs reveal accuracies ranging from 98.2\% (in case of Azure) to 38.1\% (in case of Face++), with a large disparity between males and females in the Global South (max difference of 38.5\% in case of Face++). Biases are also observed in all FRSs between females of the Global North and South (max difference of $\approx$ 50\%). A Grad-CAM analysis shows that the nose, forehead and mouth are the regions of interest for one of the open-source FRSs.  Based on this crucial observation, we design simple, low-resource bias mitigation solutions using few-shot and novel contrastive learning techniques that demonstrate a significant improvement in accuracy with disparity between males and females reducing from 50\% to 1.5\% in one of the settings. For the red-teaming experiment using the open-source Deepface model we observe that simple fine-tuning is not very useful while contrastive learning brings steady benefits.\footnote{\textcolor{red}{This work has been accepted for publication at AAAI/ACM AIES 2024.}}
\end{abstract}

\section{Introduction}
\label{sec:intro}

Artificial Intelligence (AI) systems are being developed and deployed at an unprecedented rate around the world for various applications ranging from face recognition~\cite{facepp,aws_rekognition} to web search~\cite{google_search} and chatbots~\cite{chatgpt}. Often, the development is done in a single place but the deployment is done worldwide. For example, facial recognition system AWS Rekognition~\cite{aws_rekognition} has been developed by Amazon in the USA but is deployed in other countries for tasks like ID verification~\cite{sur2021_digiyatra} at airports, with no clear distinction between different geographical deployments. Thus, if the AI model is designed without considering the deployment context, the system may end up propagating stereotypes~\cite{olier2022stereotypes} and biases~\cite{buolamwini2018gender}. 

\noindent \textbf{Facial recognition systems}: Face recognition is a group of classification tasks that involve detecting a face in an image followed by downstream tasks like gender/age/emotion detection and/or matching the input face against images stored in a database (face identification). Facial recognition systems (FRSs) are increasingly deployed for highly sensitive applications like surveillance~\cite{kamgar2011toward}, person re-identification~\cite{rao2019learning}, policing~\cite{krueckeberg2018face}, etc. This has put FRSs under the scanner of researchers and policymakers alike~\cite{axios2023,dizikes2023}.

\noindent \textbf{Impact on deployment in the Global South}: FRSs have historically been developed in countries which are defined by the UN as economically developed nations, i.e., the Global North~\cite{un_gn_gs} that includes Western countries like USA, UK and Germany, and others like Australia, Japan and Korea. A majority of the large-scale training datasets also consist of faces from this socio-economic region~\cite{raji2020saving,ma2015chicago,ma2020chicago,zhifei2017cvpr,yang2016wider,karkkainenfairface,Parkhi15,rothe2018deep,eidinger2014age}. Due to economical pricing, and relaxed licensing rules, commercial~\cite{aws_rekognition,facepp,microsoft_face} and open-source systems~\cite{Parkhi15,libfaceid,serengil2021lightface} trained on such datasets are available all around the world for use by individuals, governments and corporations alike. Multiple studies have reported large-scale biases in these platforms against minority races and genders in the Global North~\cite{buolamwini2018gender,raji2020saving,jaiswal2022two,dooley2022robustness}. The US and EU have recently strengthened their rules and introduced various measures to ensure fair and proper use of such AI platforms~\cite{nist_frt,eu_ai_act}, but the Global South countries are still lagging behind in terms of such interventions. Thus, unrestricted deployment of FRSs in the Global South may have more serious ramifications like denial of services or facilities~\cite{mit_uber_india}, without legal recourse and impacting individuals' quality of life. 

\begin{figure}[!t]
	\centering
	\begin{subfigure}{0.23\columnwidth}
		\includegraphics[width= \textwidth, height=2.5cm]{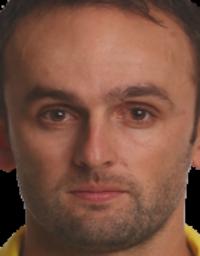}
		% \label{fig1:a}
	\end{subfigure}
	\begin{subfigure}{0.23\columnwidth}
		\includegraphics[width= \textwidth, height=2.5cm]{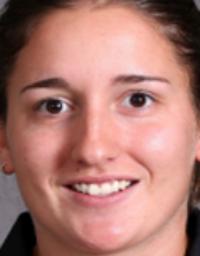}
		% \label{fig1:a}
	\end{subfigure}
     \begin{subfigure}{0.23\columnwidth}
		\centering
		\includegraphics[width= \textwidth, height=2.5cm]{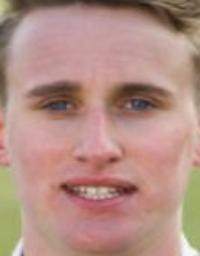}
		% \label{fig1:b}
	\end{subfigure}
	\begin{subfigure}{0.23\columnwidth}
		\centering
		\includegraphics[width= \textwidth, height=2.5cm]{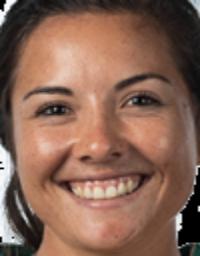}
		% \label{fig1:c}
	\end{subfigure}

    \begin{subfigure}{0.23\columnwidth}
		\includegraphics[width= \textwidth, height=2.5cm]{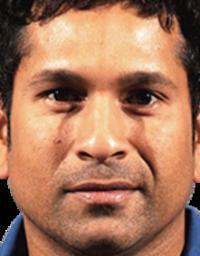}
		% \label{fig1:a}
	\end{subfigure}
	\begin{subfigure}{0.23\columnwidth}
		\includegraphics[width= \textwidth, height=2.5cm]{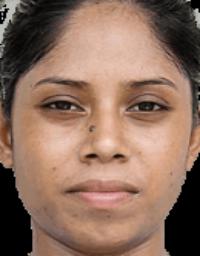}
		% \label{fig1:a}
	\end{subfigure}
     \begin{subfigure}{0.23\columnwidth}
		\centering
		\includegraphics[width= \textwidth, height=2.5cm]{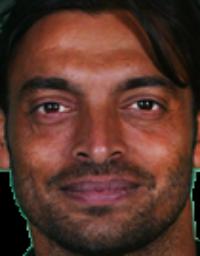}
		% \label{fig1:b}
	\end{subfigure}
	\begin{subfigure}{0.23\columnwidth}
		\centering
		\includegraphics[width= \textwidth, height=2.5cm]{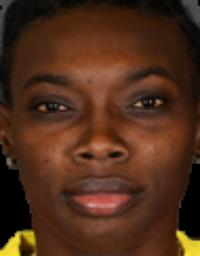}
		% \label{fig1:c}
	\end{subfigure}

    \begin{subfigure}{0.23\columnwidth}
		\includegraphics[width= \textwidth, height=2.5cm]{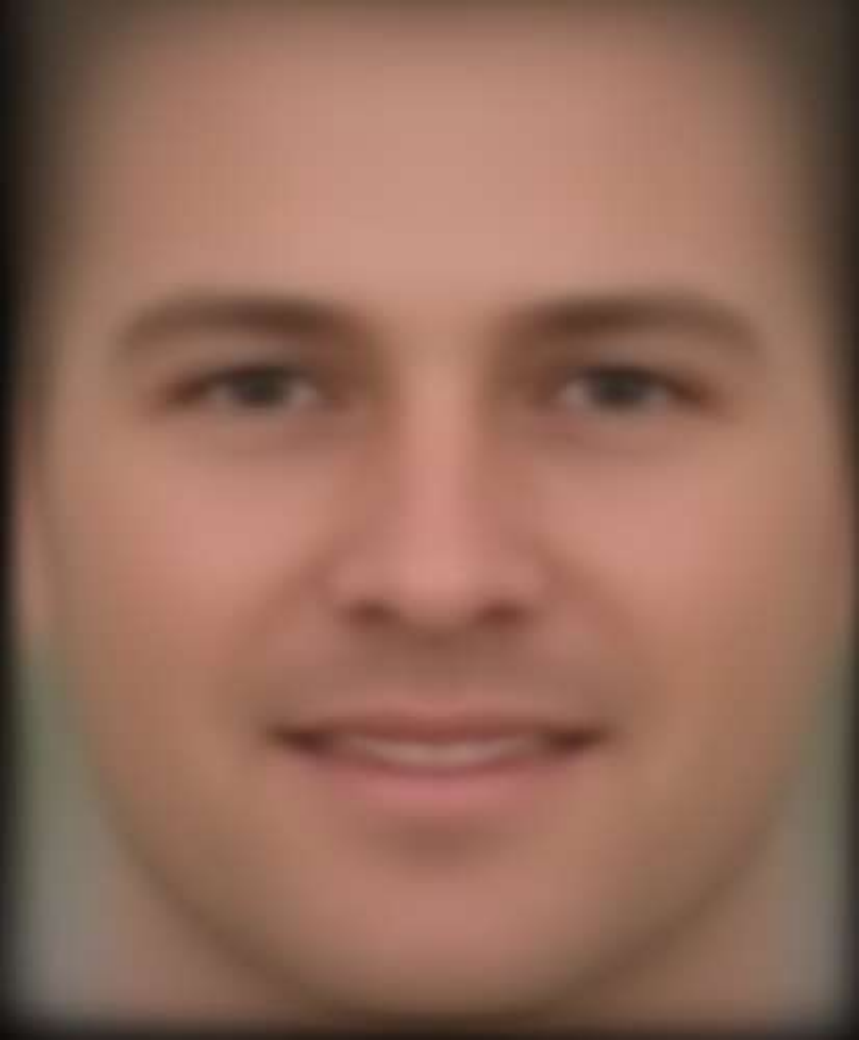}
		% \label{fig1:a}
	\end{subfigure}
	\begin{subfigure}{0.23\columnwidth}
		\includegraphics[width= \textwidth, height=2.5cm]{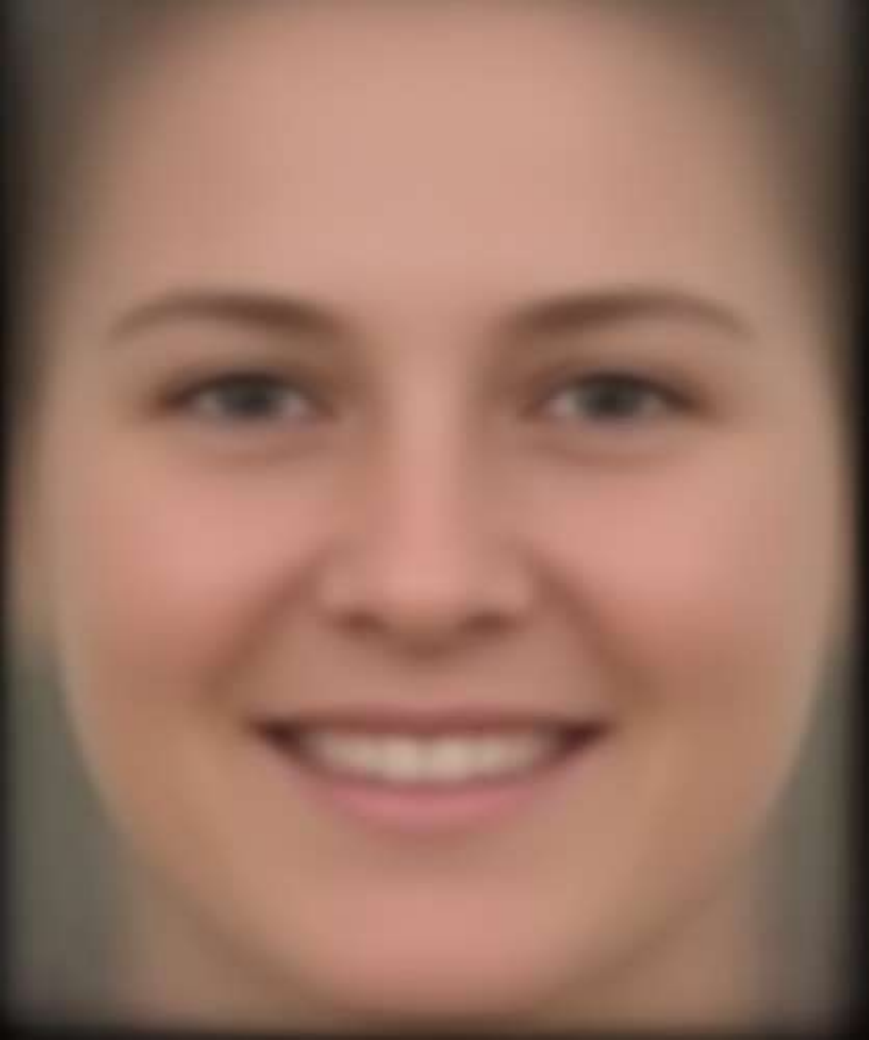}
		% \label{fig1:a}
	\end{subfigure}
     \begin{subfigure}{0.23\columnwidth}
		\centering
		\includegraphics[width= \textwidth, height=2.5cm]{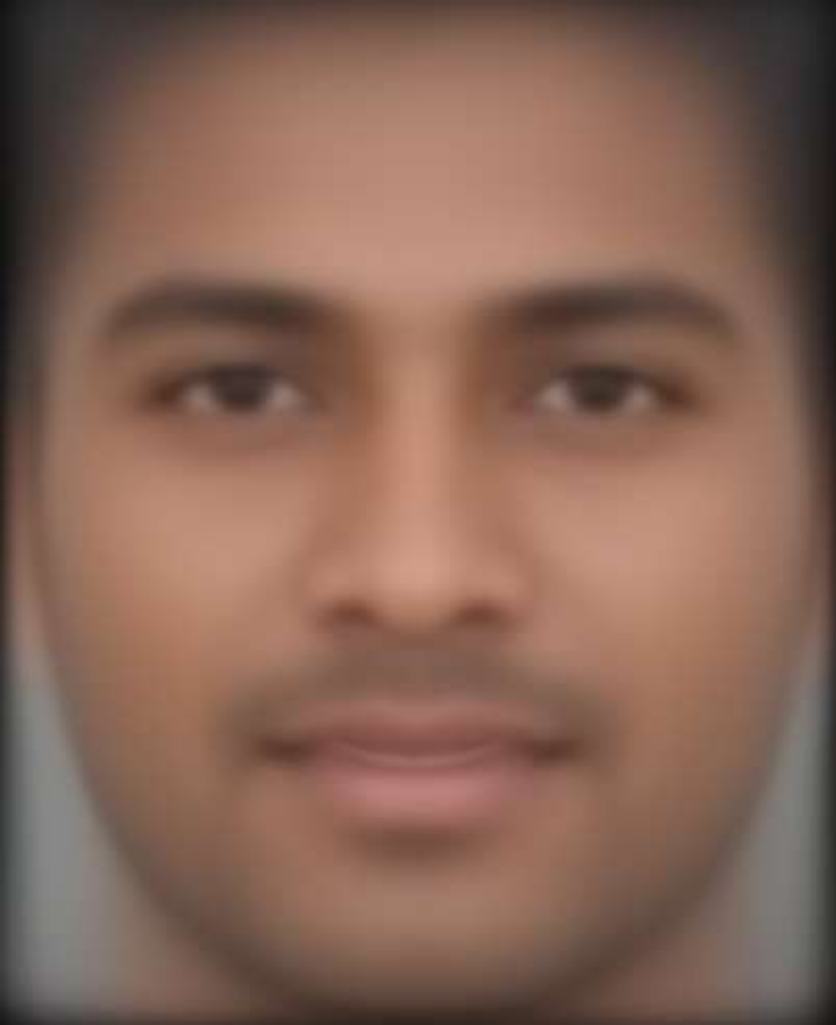}
		% \label{fig1:b}
	\end{subfigure}
	\begin{subfigure}{0.23\columnwidth}
		\centering
		\includegraphics[width= \textwidth, height=2.5cm]{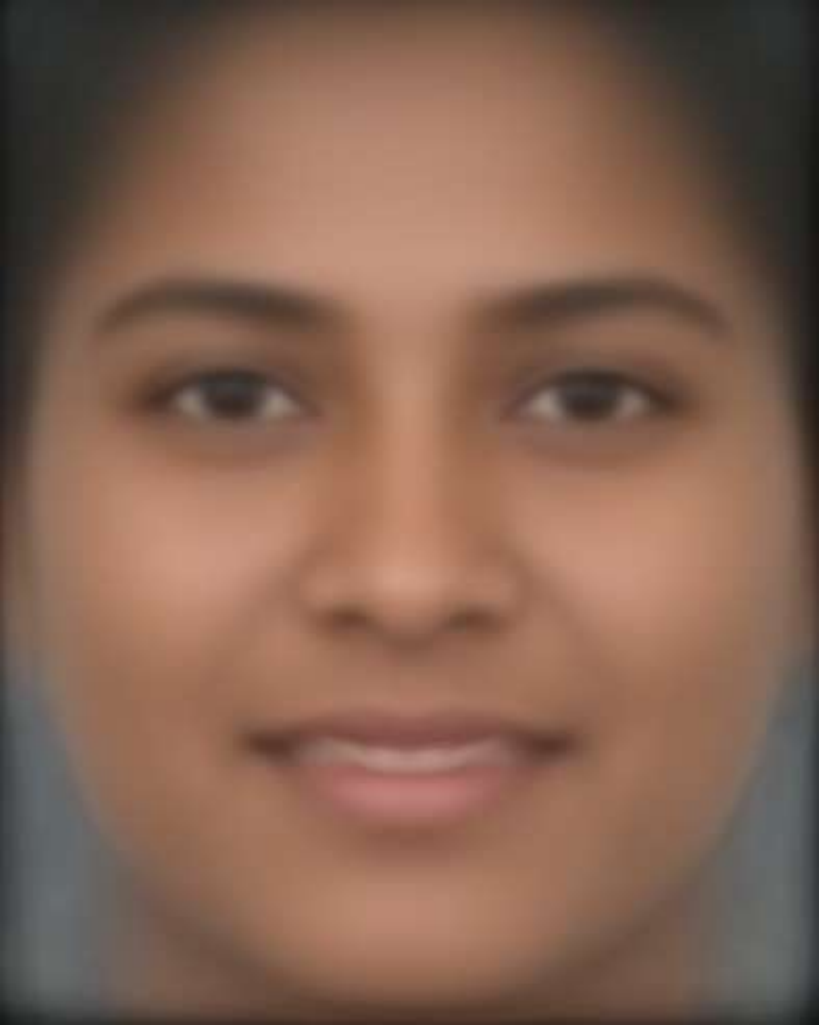}
		% \label{fig1:c}
	\end{subfigure}
	\caption{\footnotesize \textbf{Images from our \textsc{FARFace} dataset. The first row has images from Global North -- Australia, New Zealand, England and South Africa. The second row has images from the Global South -- India, Bangladesh, Pakistan and West Indies. The third row shows the average face for each region -- Global North male, Global North female, Global South male and Global South female, generated by superimposing the images of individuals from each region.}}
	\label{fig:farface_example}
\end{figure}

\noindent \textbf{A face dataset from the Global South}: Developing and benchmarking any deep learning model like FRSs, requires a large amount of well-annotated data. Unfortunately, there is a lack of such face image data from the Global South, be it South Asia or South America. To evaluate and improve model efficiency such datasets are very important. This brings us to the first research gap in literature-- \textit{Unavailability of well-annotated data for faces from the Global South.} Moreover, the existing face datasets do not have adversarial realistic variants for real-world training. Prior research~\cite{jaiswal2022two,dooley2022robustness} has shown that FRS biases increase manifold for such adversarial input. Thus, the second gap that we identify is-- \textit{Unavailability of realistic adversarial variants for face images.} While FRSs may perform well on standard color images, they may be confounded on realistic adversarial inputs that can impact real-world deployment. To bridge the gaps stated above, we present a new inclusive, adversarial and robust dataset -- \textbf{\textsc{FARFace}}, of cricketers (famous sports individuals, see Figure~\ref{fig:farface_example} for some examples), which is composed of more than 50\% individuals from the Global South. This dataset allows for \underline{f}air, \underline{a}dversarial and \underline{r}obust \underline{face} recognition model training (as our experimental results indicate later). 

\noindent \textbf{Benchmarking FRSs for the Global South}: The first step toward evaluating the efficacy of any AI model is to benchmark it for a specific deployment scenario. In this work, we benchmark FRS models for the task of \textit{gender prediction}. This is one of the most commonly deployed applications of FRSs, used for personalized recommendations~\cite{chen2007content,dai2009personalized}, surveillance, assistive technologies, safety, and marketing~\cite{ali2019discrimination}. Some of these use cases are sensitive, and any discriminatory performance can have a damaging impact on the targets~\cite{buolamwini2018gender,raji2020saving,jaiswal2022two}. Since the FRSs are often developed in one region (geographical or socio-economic) and deployed in many, without any target-specific modifications, it is important to benchmark these models' performance, especially in the Global South, where aggrieved individuals may have fewer options for recourse, legal or otherwise. This brings us to our first research question, \textbf{(RQ1.)} 
\textit{How do existing Facial Recognition Systems perform on the \textsc{FARFace} dataset, which has more than half of the faces from the Global South?}

\noindent \textbf{Explaining predictions of FRSs}: As a next step, we attempted to understand how a model converges on its prediction. Note that this question is relevant only for the open-source models, where the internal operations are known.
We use a state-of-the-art explainability tool for computer vision tasks -- Grad-CAM~\cite{selvaraju2017grad}, to identify the regions of interest in the images for the Deepface FRS by analysing the activation maps. This gives us \textbf{(RQ2.)} \textit{Why do the misclassifications occur for different faces in the case of open-source models?} A suitable answer to this question could play a key role in designing solutions that can reduce the number of misclassifications and, hence, the disparity.

\noindent \textbf{Mitigating biases in FRSs for the Global South}: Despite multiple reports on biases in FRSs for sensitive tasks like gender prediction, commercial vendors continue to develop and release these models for public use. As these black-box models cannot be debiased, researchers and activists have resorted to either calling for a complete ban of FRS usage~\cite{ban_frs} or various bias mitigation measures on open-source models. It is already known that training a deep model from scratch is extremely data and resource intensive and incurs a significant energy cost impacting the environment, hence re-training models may not be a good technique. It is beneficial to fine-tune existing pre-trained models to adapt them for a particular task and domain. As already pointed out, there is a significant lack of proper large-scale datasets from the Global South, a gap we aim to address with our work. We use low-resource, low-data and smart techniques like few-shot and contrastive learning to improve the efficiency of the Deepface FRS, with a specific focus on reducing the reported bias against individuals from the Global South. To perform this, we utilize our FARFace dataset, a more inclusive, complementary dataset to existing ones, with all its adversarial variants. Thus, our final research question is \textbf{(RQ3.)}~\textit{Can existing biases and lack of robustness in open-source FRS models be mitigated, leading to an overall improved performance, using simple interventions?} 

\noindent \textbf{Our contributions}: Here, we present a new large-scale, geographically diverse and adversarial face dataset \textsc{FARFace} and benchmark it on \textit{five} facial recognition systems -- three commercial ones viz. Amazon AWS Rekognition~\cite{aws_rekognition}, Microsoft Azure Face~\cite{microsoft_face} and Face++~\cite{facepp}, and two open-source ones viz. DeepFace~\cite{serengil2021lightface} and Libfaceid~\cite{libfaceid} -- for the gender prediction task, for both the standard condition (normal face images), as well as for various adversarial conditions -- noisy filters of RGB, Greyscale, Spread, and an occlusion filter of face mask.  
We observe that the gender prediction accuracy for females in the Global South is worse consistently across the different FRSs. To further understand the reasons behind the misclassifications (for the open-source models), we use the Grad-CAM explainability tool and analyse its activation maps. 
Finally, we adopt two different paradigms, (a)~few shot learning and (b)~contrastive learning, to mitigate the biases observed in the open source FRS. We show that our fine-tuned model also transfers well to other datasets from the Global North, thus being equally useful in that region. 
We also perform an alternative red-teaming experiment with the Deepface model to expose biases in tasks other than gender prediction-- we use the Deepface model for the task of country prediction (and by extensions, ethnicity) from face images~\cite{batsukh2016effective,albdairi2020identifying}, especially for individuals from the Global South. The results from this experiment reinforce the cautionary message that such tasks, while easy to design, are unfair and biased and should not be propagated.
The primary insights from our work are as follows:

\noindent
$\bullet$ Our extensive experiments show that the disparity in accuracy between males and females for gender prediction is lower in the Global North (maximum disparity of 10\% for commercial FRS-- Face++ and 66.77\% for open-source FRS-- Libfaceid) than in the Global South (maximum disparity 38.51\% for commercial FRS-- Face++ and 83.35\% for open-source FRS-- Libfaceid). 

\noindent
$\bullet$ There is a consistent trend of high disparity for adversarial inputs of females from Global South (similar to~\citet{jaiswal2022two}), indicating a lack of model robustness. Grad-CAM analysis shows systematic regions of interest for male classification, whereas they are random for females. 

\noindent
$\bullet$ Finally, our proposed approaches to overcome the gender disparity in FRSs (for gender prediction task) by adopting few-shot and contrastive learning demonstrate improvements in female accuracy by 59\% and 60\%, respectively.

\noindent
$\bullet$ As an additional red-teaming task, we also study the performance of FRSs for predicting the country from the face. Our experiment shows that such a task is highly biased and can lead to unfavourable outcomes for individuals from both the Global North and South. Simple fine-tuning is not very useful; however, we get benefits in both accuracy and disparity reduction when contrastive learning is effectively used.

\noindent \textbf{Dataset availability}:
The \textsc{FARFace} dataset is available for research purposes, on request\footnote{Dataset Request: \url{https://forms.gle/2Nd9ntNcc71vvwEJ7}}.
We would like to state here that the original dataset collected by us is gender imbalanced (86\% males). This is reflective of the gender imbalance in society, including sports~\cite{toi_puru_2023}. To prevent further misuse, we are releasing only a balanced subset (with all corresponding adversarial variants). 

\section{Background \& related work}
\label{sec:relwork}
We now present a brief overview of the literature on face datasets and bias mitigation in FRSs.

\noindent \textbf{Face datasets}: 
There exist a multitude of face datasets in literature meant for different face recognition tasks. Some are designed specifically for audit studies~\cite{buolamwini2018gender} and could be balanced~\cite{raji2020saving}, be collected from volunteers~\cite{ma2015chicago,ma2020chicago,lakshmi2021india}, have multiple age~\cite{zhifei2017cvpr,eidinger2014age,Parkhi15} and racial~\cite{rothe2018deep,zhifei2017cvpr,karkkainenfairface} groups. All these datasets have images mainly of individuals from the Global North and have no adversarial images for robust model training or testing. 
We \textit{address both these gaps} through our \textsc{FARFace} dataset.

\noindent \textbf{Audit \& bias mitigation on FRSs}: 
Audits on commercial FRSs~\cite{buolamwini2018gender,raji2020saving,jaiswal2022two} and open-source FRSs~\cite{dooley2022robustness}, have shown temporal persistence to biases. Adversarial audits~\cite{jaiswal2022two,dooley2022robustness,majumdar2021unravelling} on existing datasets show that disparities get magnified for noisy inputs. Our audit lies at the intersection of~\cite{jaiswal2022two} and~\cite{dooley2022robustness,majumdar2021unravelling} -- we audit both commercial and open-source FRSs for the task of gender prediction. Similar to~\citet{majumdar2021unravelling}, we also study the activation maps
for the prediction outputs but for the task of gender prediction instead of landmark detection. 
Another prior work by~\citet{wang2022fairness} proposes bias mitigation strategies by developing adversarial filters on input images in real time. However, our approach differs from the mentioned approach (FAAP~\cite{wang2022fairness}) in the following ways: (i)~Their adversarial filters do not simulate any real-world situation, and the authors do not disclose what type of filter is developed, hence it is difficult to compare against or audit for; (ii)~Their adversarial filters conceal the gender and race information and improve performance for tasks like \textit{smile} or \textit{hair color} identification, whereas we explicitly audit the platforms for gender prediction, which has a large number of downstream applications ranging from marketing to surveillance.
Furthermore, most existing approaches are either too complex or need significant changes to the pipeline~\cite{gong2020jointly,wang2020mitigating,wang2021meta,conti2022mitigating,wang2022fairness}. We, on the other hand, perform simple fine-tuning and contrastive learning to mitigate the observed biases.

Thus, in this work, we perform all three parts of a Responsible AI pipeline for FRSs viz. (i)~benchmark audits of commercial and open-source FRSs on a large scale inclusive, adversarial dataset for gender bias (if any), 
(ii)~explaining model predictions to better understand the reason behind misclassifications of images of several demographics and, (iii)~mitigation of the said biases.
\section{\textsc{FARFace} Dataset}
\label{sec:dataset}
In this section, we describe the collection \& preparation process of the \textsc{FARFace} dataset followed by its basic demographic distribution based on \textit{gender} and \textit{geographic region}.

\begin{figure}[!t]
	\centering
 	\begin{subfigure}{0.18\columnwidth}
		\centering
		\includegraphics[width= \textwidth, height=1.75cm]{images/dataset/Shamilia_Connell_0.jpg}
	    \caption{\tiny ORIG}	
        \label{fig:adv:orig}
	\end{subfigure}
     \begin{subfigure}{0.18\columnwidth}
		\includegraphics[width= \textwidth, height=1.75cm]{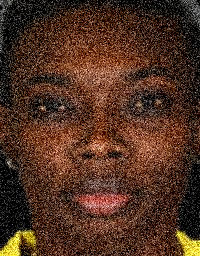}
        \caption{\tiny RGB$_{0.3}$}
		\label{fig:adv:rgb}
	\end{subfigure}
	\begin{subfigure}{0.18\columnwidth}
		\includegraphics[width= \textwidth, height=1.75cm]{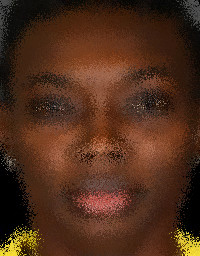}
        \caption{\tiny Spread}
		\label{fig:adv:spread}
	\end{subfigure}
     \begin{subfigure}{0.18\columnwidth}
		\centering
		\includegraphics[width= \textwidth, height=1.75cm]{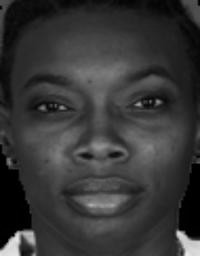}
	    \caption{\tiny Greyscale}	
        \label{fig:adv:grey}
	\end{subfigure}
	\begin{subfigure}{0.18\columnwidth}
		\centering
		\includegraphics[width= \textwidth, height=1.75cm]{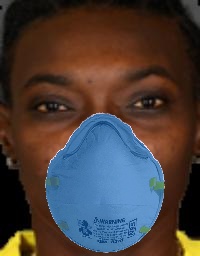}
	    \caption{\tiny N95 mask}	
        \label{fig:adv:mask}
	\end{subfigure}
	\caption{\footnotesize \textbf{Adversarial variants in the \textsc{FARFace} dataset, shown for an example image (original image in (a)).}}
	\label{fig:farface_adversarial}
\end{figure}

\subsection{Description of the dataset}
\textsc{FARFace} is curated from the face images of male and female cricket players belonging to eight prominent cricket-playing countries around the world -- Australia, New Zealand, England, South Africa (Global North) and, Bangladesh, Pakistan, India and West Indies (Global South). We collect these images by scraping the \textsc{ESPNCricInfo} website's player pages\footnote{https://www.espncricinfo.com/cricketers} for \textit{all} domestic and international cricketers of the above countries. 
While there are multiple other cricketing nations all around the world, our choice was guided by the following reasons-- (i)~most players in the Global North countries have lighter skin tones, whereas a majority of the players in the Global South countries have darker skin tones, giving us more diversity in terms of skin tone (see Fig.~\ref{fig:farface_example}), face structure and geography, and (ii)~these countries have both male and female cricketers, giving us gender diversity. 
Table~\ref{tab:dataset_stats} presents the number of images, and gender distribution for all the countries.

We next describe the methodology of dataset collection and generating their adversarial variants. 

\begin{table*}[!t]
	\scriptsize
	\begin{center}
		\begin{tabular}{| c | c | c | c || c | c || c | c || c |}
			\hline
			\multirow{3}{*}{\textbf{Region}} & \multirow{3}{*}{\textbf{Country}} & \multicolumn{6}{c|}{\textbf{Type of Image}} & \multirow{3}{*}{\textbf{Total (A$\times$4 + B + C)}}\\ 
			\cline{3-8}
			& & \multicolumn{2}{c||}{\textbf{ORIG / RGB$_{0.3}$~/ RGB$_{0.5}$~/ SPRD (A)}} & \multicolumn{2}{c||}{\textbf{GREYSCALE (B)}} & \multicolumn{2}{c||}{\textbf{MASKED (C)}} & \\
			\cline{3-8}
			& & \textbf{Male} & \textbf{Female} & \textbf{Male} & \textbf{Female} & \textbf{Male} & \textbf{Female} & \\
			\hline\hline
			\multirow{4}{*}{\textbf{Global North}} & Australia & 666	& 159  & 759 & 181 & 651 & 158 & 5,049\\
			\cline{2-9}
			& New Zealand & 322 & 232 & 365 & 232 & 312 & 221 & 3,346\\
			\cline{2-9}
			& England & 947 & 112 & 1,276 & 133 & 917 & 111 & 6,673\\
			\cline{2-9}
			& South Africa & 444 & 59 & 482 & 59 & 422 & 58 & 3,033\\
			\hline\hline
			\multirow{4}{*}{\textbf{Global South}} & Bangladesh & 328 & 39 & 328 & 39 & 319 & 39 & 2,193\\
			\cline{2-9}
			& India & 1,966 & 187 & 2,066 & 188 & 1,966 & 187 & 13,019\\
			\cline{2-9}
			& Pakistan & 442 & 79 & 478 & 79 & 435 & 78 & 3,154\\
			\cline{2-9}
			& West Indies & 533 & 64 & 584 & 65 & 488 & 63 & 3,588\\
			\hline\hline
			\multicolumn{2}{|c|}{\textbf{All}}  & \textbf{5,648} & \textbf{931} & \textbf{6,338} & \textbf{976} & \textbf{5,510} & \textbf{915} & \textbf{40,055}\\
			\hline
		\end{tabular}
		\caption{\footnotesize \bf Distribution of images across countries, gender and image type in the \textsc{FARFace} dataset. A majority of the dataset is composed of images from the Global South countries (54.8\%), and has male individuals (85.98\%). Among all the types, greyscale type images have the largest share (18.25\%).}
		\label{tab:dataset_stats}
	\end{center}
\end{table*}

\subsection{Dataset curation}
To collect the dataset, we use the Selenium tool to collect the images and the player metadata like country, name and gender. We perform the following steps to clean and preprocess the images:
 
 \noindent
 $\bullet$ We remove all images which have drawings instead of photographs. We also set aside all images that are greyscale by default (these are added back later). 

 \noindent
 $\bullet$ We crop and resize all resulting images to display only the face area using YOLOv5~\cite{glenn_jocher_yolov5}. The resized images have a resolution of 200$\times$256px. We choose this resolution heuristically as it creates the least distortion of the images. YOLOv5 also provides the bounding box coordinates of the final faces. We will release only the cropped images themselves. 

We are finally left with 5,648 male images and 931 female images, all in JPEG format. We henceforth refer to this set of original images as the ORIG set. 

\noindent \textbf{Dataset accuracy}:
We manually checked the country, name and gender for a randomly chosen 15\% subset of the dataset, and matched the data with the image. The information was accurate in all cases, showing the sanctity of the dataset.

\noindent \textbf{Adversarial variants}: After the above ORIG set of images was prepared, we used the GIMP\footnote{https://www.gimp.org/} image editor to create the following adversarial variants -- 
RGB (with noise values 0.3 and 0.5), Spread and Greyscale. 
RGB and Spread have been used previously by \citet{jaiswal2022two} to create realistic adversarial inputs for auditing FRSs. RGB simulates the effect of edited images from social media, and Spread simulates the effect of blurry photos captured by cameras, which may be exposed to natural elements like rainwater. We introduce Greyscale for the first time to simulate the effect of a black and white photograph, thus removing the skin-tone of the individual in the photograph. 
We also use a popular tool MaskTheFace~\cite{anwar2020masked} to create \textit{masked} variants of the ORIG set by applying blue N95 face masks on the images, to create an occlusion-based adversary. Example images for each of the adversarial variants are shown in Figure~\ref{fig:farface_adversarial}.

\subsection{Basic statistics}
Our downloaded set has 7,324 images, which reduces to 6,579 after cleaning. In the cleaned dataset, 85.85\% images are of males, and 14.15\% are of females. 
The size of the two RGB sets and the Spread set are the same as the ORIG set. The Greyscale set is slightly larger as it also includes some images which were greyscale by default. The Masked set is slightly smaller because the masking tool did not identify faces in all images. Thus, we effectively get 40,055 images in the \textsc{FARFace} dataset after generating the five adversarial variants from the original set of images. The gender distribution in this final dataset is 85.98\% males and 14.02\% females.
Some sample images are present in Figure~\ref{fig:farface_example}.  

\noindent \textbf{Task description}: While FRSs can be used for multiple tasks like face detection, gender/age detection and identification, in this paper, we only perform gender detection from the input face since we have the ground truth data for only this task. Even though the player profiles are labelled with age, we do not experiment for it as there is no guarantee on the age being the same as when the photographs were clicked. Further, a manual inspection reveals that a majority of the dataset is of young people (expected of sportspersons) and may cause a model to perform poorly when predicting the age for old people. We acknowledge that gender is a spectrum, but in this work, we consider it to be binary because all FRS models only predict either male or female as a label for the task of gender prediction~\cite{keyes2018misgendering}. The ground truth labels for all individuals in our dataset are also only male or female. We also do not study face verification/identification because we only have 1 image per identity. 

As an additional red-teaming task, we also perform country prediction from the input face for one of the open-source FRSs. The results show that such tasks are not only inaccurate but also highly biased.
\section{FRS platforms \& Audit methodology}
\label{sec:platform}

We now give a brief overview of the Face Recognition tools/platforms we audit using our benchmark dataset. Next we describe the methodology of our audits, Grad-CAM analysis and bias mitigation on one of the classical and highly accurate face detection models-- VGG-Face~\cite{Parkhi15}. We use the implementation in Deepface~\cite{serengil2021lightface}\footnote{\label{deepface_footnote}https://github.com/serengil/deepface}, which is a modified version of VGG-Face. 

\subsection{Platforms audited}
In this study, we audit five FRSs-- three popular, economically-priced plug-and-play API-based commercial FRSs and two popular pre-trained open-source FRSs.

\noindent \textbf{Commercial FRSs}: 
We perform our audits on Amazon AWS Rekognition~\cite{aws_rekognition}, Microsoft Azure Face~\cite{microsoft_face} and Face++~\cite{facepp}. These models are made available through easy-to-use APIs and charge nominal fees for their services; thus are easy to deploy for non-domain experts at scale. However, no information is shared publicly on the training dataset or the model architecture, making it impossible for third-party researchers to address any performance or bias issues. Multiple previous audits~\cite{buolamwini2018gender,raji2020saving,jaiswal2022two} have exposed large-scale biases in these platforms for standard benchmark datasets.

\noindent \textbf{Open-source FRSs}: We audit two popular pre-trained open-source models-- Libfaceid~\cite{libfaceid}\footnote{https://github.com/richmondu/libfaceid} (a modified version of CaffeNet~\cite{krizhevsky2012imagenet}) and Deepface~\cite{serengil2021lightface}\footnotemark[\getrefnumber{deepface_footnote}] (implementation of VGG-Face~\cite{Parkhi15}). The training data information for Libfaceid is not available; Deepface has been pre-trained on two datasets-- VGG-Face for the initial face recognition task and WikiData (from IMDB-Wiki~\cite{rothe2018deep}) for the gender prediction task. Both these datasets have $> 50\%$ faces of White individuals (primarily found in Global North countries). As there are no licensing restrictions on their deployment, these models can propagate their pre-training biases, if deployed in geographies which are not represented well during training. These models are available for download and use, but require domain knowledge for deployment. They can be fine-tuned and modified for specific use-cases allowing improvement of performance and reducing observed biases. We can extract output explanations to better understand the model's decisions. 
Both employ standard CNN-based architectures with Libfaceid having 11 layers (3 convolution layers) and Deepface having 22 layers (13 convolution layers); Deepface is a deeper model and captures more detailed facial features compared to Libfaceid. 

\subsection{Methodology of experiments}
In this work, we perform three sets of experiments described as follows. 

\noindent
$\bullet$ \textbf{Audit of all FRSs}: We audit all five FRSs on the entire \textsc{FARFace} dataset for the task of gender prediction. Each FRS is supplied with raw image files in JPEG format and the response for gender prediction is collected.

\noindent
$\bullet$ \textbf{Grad-CAM analysis of Deepface}: We analyse the predictions for the Deepface model using the Grad-CAM explainability tool to better understand the correct and incorrect predictions for the two socio-political regions and genders.
    
\noindent
$\bullet$ \textbf{Bias-mitigation of Deepface}: As has been observed in previous literature, and as we shall also show through our audits in the subsequent sections, all FRSs demonstrate biases against the females and more so in the Global South (faces of individuals with darker skin tones and different facial structures). We shall use two simple techniques to show that it is quite straightforward to mitigate the observed biases in such open-source models. We choose Deepface for this as it is a deeper model and is better designed to detect more interesting facial features as opposed to Libfaceid; thus an analysis on this model is expected be more generalizable.

\noindent \textbf{(a) Few-shot fine-tuning}: We perform one-shot and two-shot fine-tuning on the pre-trained Deepface model. To choose our one-shot data points, we randomly sample one image from each country and gender, thereby giving us 16 unique data points; similarly 32 unique data points are chosen for the two-shot scenario. For each learning setup, we have three settings -- involving 0\%, 50\% and 100\% adversarial examples from the RGB$_{0.3}$ set. In our experiments we find the RGB filter to be most adversarial resulting in the highest performance drop (see Figure~\ref{fig:benchmark_audit}), and hence the choice. This allows us to target two objectives-- reducing observed gender and demographic bias and improving model robustness to adversarial inputs. The hyperparameters are -- $\alpha: 1e{-5}$, epochs: 10, optimizer: ADAM.
  
\noindent \textbf{(b) Contrastive learning}: Contrastive learning has proved to be very effective in handling adversarial inputs~\cite{hateproof}. In this paper we take dual advantage of the contrastive learning setup. While on one hand, we teach the model the contrast between males and females, on the other hand we also teach it that the original and the adversarial versions of a face are the same. Based on this intuition, we propose a triplet loss~\cite{schroff2015facenet} function to perform a \textit{novel} contrastive learning based fine-tuning. The anchor points (any data point in a given class) ($x_a$) are chosen from each gender and country, the positive example for each anchor ($x_{a}^{+}$) is an adversarial image of the same individual, whereas the negative example ($x_{a}^{-}$) is an image of another person from the ORIG set of the opposite gender. We choose the ($x_{a}^{+}$) from RGB$_{0.3}$ in this case to improve  model robustness while reducing bias simultaneously and we choose the ($x_{a}^{-}$) from the opposite gender to create a stronger negative example for every anchor. We also experiment with a setup where the negative example is chosen from the same or opposite gender with a probability (more results are shared in the Supplementary section).
The hyperparameters are -- $\alpha: 1e{-5}$, epochs: 40, optimizer: ADAM, $L_{triplet} : L_{BCE}$ = 0.8:0.2. 

\noindent\textbf{Held-out test set for all experiments}: A held out test set of 480 images is created, composed of 60 images from each country, in a 2:1 ratio of males to females. Each setting is evaluated three times, and the avg. percentages are reported.
We run our experiments on an Ubuntu 18.04 LTS Intel(R) Xeon(R) Gold 6126 CPU server with NVIDIA Tesla P100 GPU (CUDA v11.4), 128 GB RAM and 48 cores. 

\section{Results \& observations}
\label{sec:results}
We now present the results for our benchmarking audit, followed by the Grad-CAM analysis, bias mitigation and red-teaming task of country prediction.

\subsection{Benchmark audit of FRSs (RQ1)} 

\begin{figure*}[!ht]
	\centering
    \begin{subfigure}{0.66\columnwidth}
		\includegraphics[width= \textwidth, height=3cm,keepaspectratio]{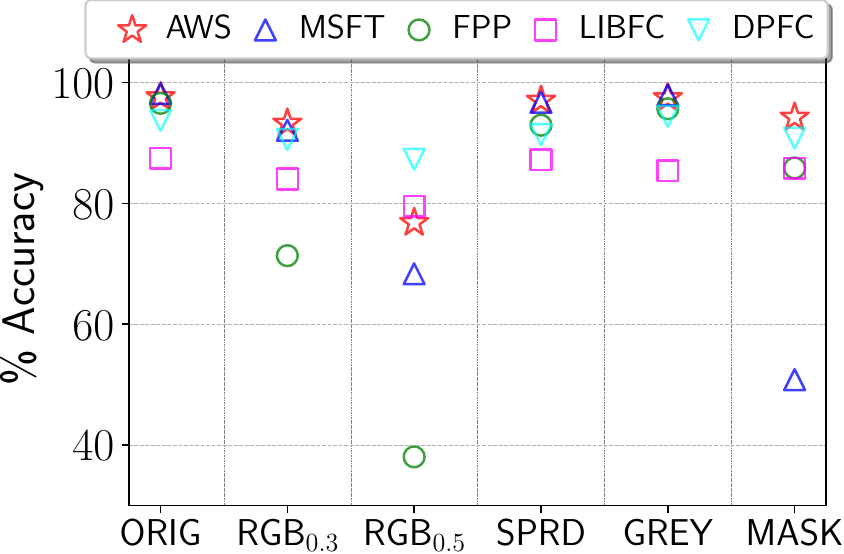}
	    \caption{Overall accuracy for all images}	
        \label{fig:overall:all}
	\end{subfigure} 
	\begin{subfigure}{0.66\columnwidth}
		\includegraphics[width= \textwidth, height=3cm,keepaspectratio]{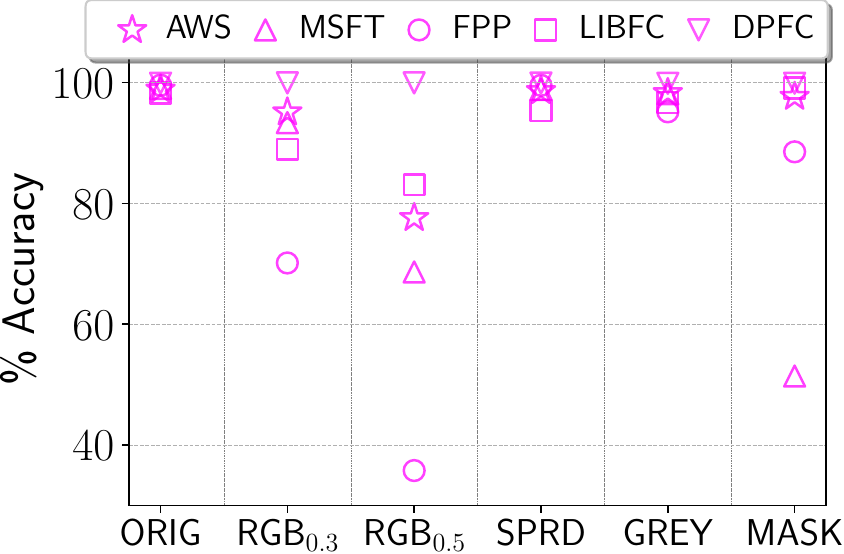}
	    \caption{Overall accuracy for males}	
        \label{fig:overall:males}
	\end{subfigure} 
	~\begin{subfigure}{0.66\columnwidth}
		\includegraphics[width= \textwidth, height=3cm,keepaspectratio]{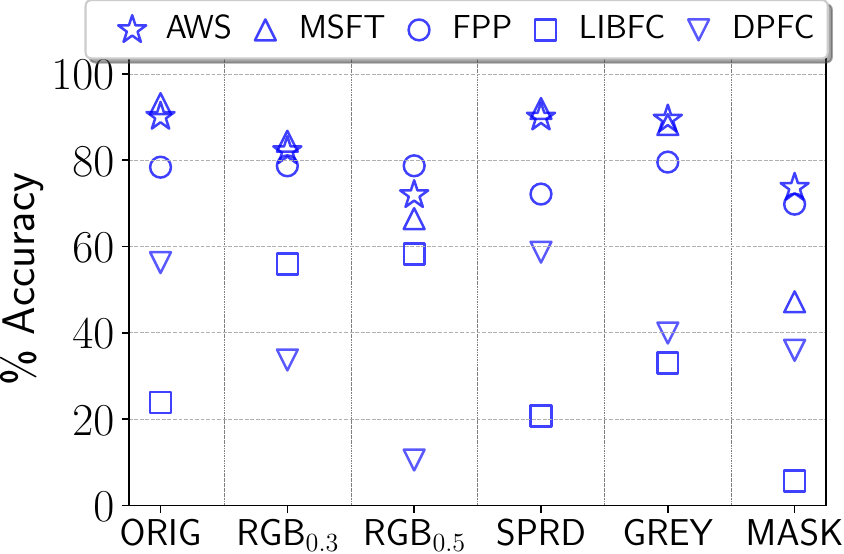}
	    \caption{Overall accuracy for females}	
        \label{fig:overall:females}
	\end{subfigure}	
	\caption{ \footnotesize \bf Overall accuracy for all FRSs, segregated by image type, for all images (a) and for each gender group (b,c). On average, AWS and Deepface are the best performing commercial and open-source FRSs respectively, independent of the gender. All the FRSs are least robust to RGB$_{0.5}$ for both genders and MASK for females. The FRSs are AWS Rekognition (AWS), Microsoft Azure Face (MSFT), Face++ (FPP), Libfaceid (LIBFC), Deepface (DPFC).}
	\label{fig:benchmark_audit}
\end{figure*}

In Figure~\ref{fig:benchmark_audit}, we show the overall accuracy and the gender-wise results of the different FRSs for the benchmark audit of \textsc{FARFace} dataset separated by image type. From Fig.~\ref{fig:overall:all}, we observe that all FRSs perform well on the original set (ORIG), with Microsoft Azure Face being the best and identifying the correct gender for 98.22\% of the images. On the other hand, Libfaceid correctly identifies the gender in 87.48\% images. All FRSs report low accuracies for the RGB variants (highest -- 93.3\% on AWS for RGB$_{0.3}$; lowest -- 38.05\% on Face++ for RGB$_{0.5}$), but are robust to the Spread and Greyscale variants (performance on both these variants is comparable to the ORIG set). For the masked variant, Azure reports the lowest accuracy of 51\%, with others reporting $> 85\%$. Among the FRSs studied in this paper, AWS Rekognition and Deepface are the best performing commercial and open-source FRSs respectively. Commercial FRSs perform better than open-source ones on five of the image sets. Deepface open-source FRS is the best performing FRS for the RGB$_{0.5}$ variant. We calculate the standard deviation of the accuracy distribution across different image sets for each FRS. In general, commercial FRSs are less stable, with the standard deviation varying from 8\% (AWS) to 22.6\% (Face++), but for open-source FRSs, this is less than 3\%.

We also perform a deeper analysis of accuracies for males (Fig.~\ref{fig:overall:males}) and females (Fig.~\ref{fig:overall:females}). From Fig.~\ref{fig:overall:males}, for gender prediction on male images, we see that all FRSs perform exceptionally well on the ORIG set, with Deepface being the best, correctly predicting the male gender for 99.91\% of the images. On the other hand, the lowest accuracy is reported by Libfaceid at 98.31\%. All FRSs except Deepface report the lowest accuracy for RGB$_{0.5}$, but are generally robust to Spread and Greyscale variants. For the masked variant, the lowest accuracy is reported by Microsoft Azure. 

Next, looking at Fig.~\ref{fig:overall:females}, we note that the trends are less systematic for female faces. 
Even on the ORIG set, the accuracies range from 90.12\% (AWS Rekognition) to 23.83\% (Libfaceid). In fact, the open-source FRSs consistently report poor accuracies (independent of the image type) with the highest being 59\% by Deepface on GREY and lowest being 5.7\% for Libfaceid on MASK. The standard deviation trends reverse here, with commercial FRSs reporting a more stable performance -- 4.14\% (Face++) to 18.18\% (Azure) as opposed to open-source FRSs which report a standard deviation of 17.52\% (Deepface) and 20.72\% (Libfaceid).

\begin{table*}[!ht]
\footnotesize
    \begin{center}
        \begin{tabular}{|c|c||c|c||c|c||c|c||c|c||c|c|}
        \hline
            \multirow{2}{*}{\textbf{Type}} & \multirow{2}{*}{\textbf{Region}} & \multicolumn{2}{c||}{\textbf{AWS}} & \multicolumn{2}{c||}{\textbf{MSFT}} & \multicolumn{2}{c||}{\textbf{FPP}} & \multicolumn{2}{c||}{\textbf{LIBFC}} & \multicolumn{2}{c|}{\textbf{DPFC}} \\ \cline{3-12}\cline{3-12}
            ~ & ~ & M & F & M & F & M & F & M & F & M & F \\ \hline\hline
            \multirow{2}{*}{ORIG} & GN & 98.95 & 95.37 & 99.24 & 96.26 & 99.33 & 89.68 & 97.02 & 30.25 & 99.92 & 75.98 \\ \cline{2-12}
            ~ & GS & 98.81 & \textbf{82.11} & 98.93 & \textbf{88.35} & 99.76 & \textbf{61.25} & 98.53 & \textbf{15.18} & 99.91 & \textbf{26.29} \\ \hline\hline
            \multirow{2}{*}{RGB$_{0.3}$} & GN & 93.57 & 88.61 & 94.12 & 91.64 & 59.77 & 85.05 & 84.95 & 65.48 & 99.92 & 47.51 \\ \cline{2-12}
            ~ & GS & 96.21 & \textbf{72.36} & 92.84 & \textbf{73.44} & 77.7 & \textbf{69.11} & 91.28 & \textbf{42.82} & 100 & \textbf{12.74} \\ \hline
        \end{tabular}
        \caption{ \footnotesize \textbf{Average male and female gender prediction accuracies for the images belonging to the ORIG and the RGB$_{0.3}$ set from Global North and Global South. M \& F refer to the male and female groups, the FRSs are AWS Rekognition (AWS), Microsoft Azure Face (MSFT), Face++ (FPP), Libfaceid (LIBFC), Deepface (DPFC). Accuracy for male images are comparable between the two regions, but that for female images are lower than males consistently across regions and FRSs. FRSs report significantly lower accuracy for female images from Global South than for male images within the same region. This accuracy is also lower than accuracy observed for female images from Global North.
       } \label{tab:regionwisegenderacc}}
    \end{center}
\end{table*}

\noindent
\textbf{Disparity in accuracy is higher for images from GS: }Table~\ref{tab:regionwisegenderacc} shows the average accuracy (micro avg. for all images in a given region) of gender prediction for male and female cricketers for each region in the original (ORIG) and RGB$_{0.3}$ sets for all the five FRSs (more results for a balanced set of males and females are shared in the Supplementary section).
In the rest of this section, we choose the RGB filter to demonstrate our results since it has the highest adversarial effect on the model performances (see Figure~\ref{fig:benchmark_audit}).
For the ORIG set, all five FRSs report high accuracy (min. 97.02\%) of gender prediction for males in both Global North (GN) and Global South (GS) countries, with three FRSs performing marginally better for males in GN than in GS. 
In contrast, the accuracy of gender prediction for images of females from Global South is significantly lower than those from Global North for all FRSs consistently. This observation corroborates the observations made in prior literature~\cite{buolamwini2018gender,raji2020saving,jaiswal2022two} regarding less accurate performance of FRSs on darker female subgroups.
Irrespective of the FRS, the gender prediction accuracy for females is always lower than males within the same region. The disparity observed for commercial FRSs ranges from nearly 3\% in MSFT to nearly 10\% in Face++ for the images of cricketers from Global North. However, this disparity increases to 10\% in MSFT and 38.51\% in Face++, respectively, for images from Global South.
The disparity in accuracy across gender is even more pronounced in the open-source FRSs. For images from Global North, the disparity ranges from 23.94\% (for Deepface) to 66.67\% (for Libfaceid). However, for images from the Global South countries, this disparity goes upwards of 70\% for both the open-source FRSs.

Looking at the RGB$_{0.3}$ set, we see that there is a drop in accuracy for all FRSs on images from the Global North, but a slight improvement in accuracy for some FRSs on Global South images. Interestingly, for the Global North images, the minimum and maximum disparity between the males and females reduce to 2.5\% and 52.41\%, indicating that the accuracy for males reduces more than that for females. For the Global South images, the max disparity now becomes 87.26\%. The disparity in accuracy between females of the two regions reduces for two of the FRSs.

\noindent \textbf{Takeaways}:
\textit{(a)~Gender prediction accuracy for females is worse than that for males for all the FRSs. This phenomenon is even more pronounced in case of open source FRSs.} 
\textit{(b)~Gender prediction accuracy for females is worse for images from Global South than for those from Global North across all FRSs.}

\subsection{Grad-CAM analysis of predictions (RQ2)}

\begin{figure}[!t]
	\centering
	\begin{subfigure}{0.23\columnwidth}
		\includegraphics[width= \textwidth, height=2.5cm]{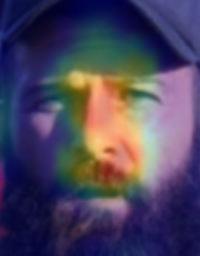}
		% \label{fig1:a}
	\end{subfigure}
     \begin{subfigure}{0.23\columnwidth}
		\centering
		\includegraphics[width= \textwidth, height=2.5cm]{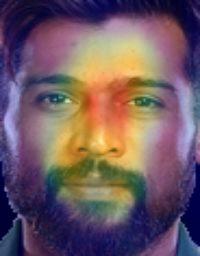}
		% \label{fig1:b}
	\end{subfigure}
	\begin{subfigure}{0.23\columnwidth}
		\centering
		\includegraphics[width= \textwidth, height=2.5cm]{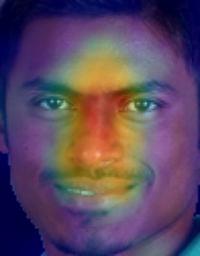}
		% \label{fig1:c}
	\end{subfigure}
    \begin{subfigure}{0.23\columnwidth}
		\centering
		\includegraphics[width= \textwidth, height=2.5cm]{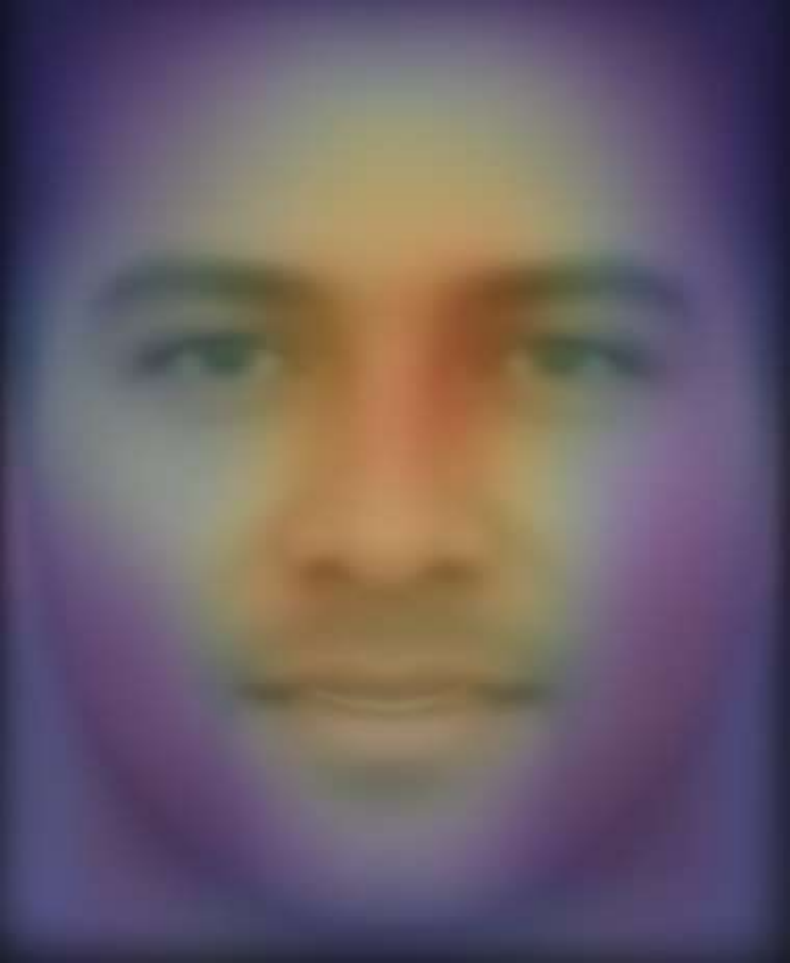}
		% \label{fig1:c}
	\end{subfigure}

	\begin{subfigure}{0.23\columnwidth}
		\includegraphics[width= \textwidth, height=2.5cm]{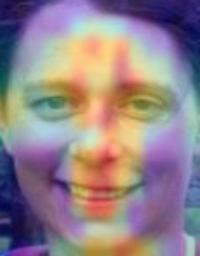}
		% \label{fig1:a}
	\end{subfigure}
     \begin{subfigure}{0.23\columnwidth}
		\centering
    	\includegraphics[width= \textwidth, height=2.5cm]{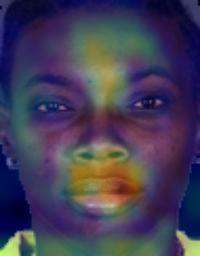}
		% \label{fig1:b}
	\end{subfigure}
	\begin{subfigure}{0.23\columnwidth}
		\centering
		\includegraphics[width= \textwidth, height=2.5cm]{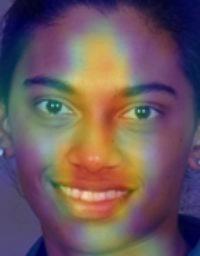}
		% \label{fig1:c}
	\end{subfigure}
    \begin{subfigure}{0.23\columnwidth}
		\centering
		\includegraphics[width= \textwidth, height=2.5cm]{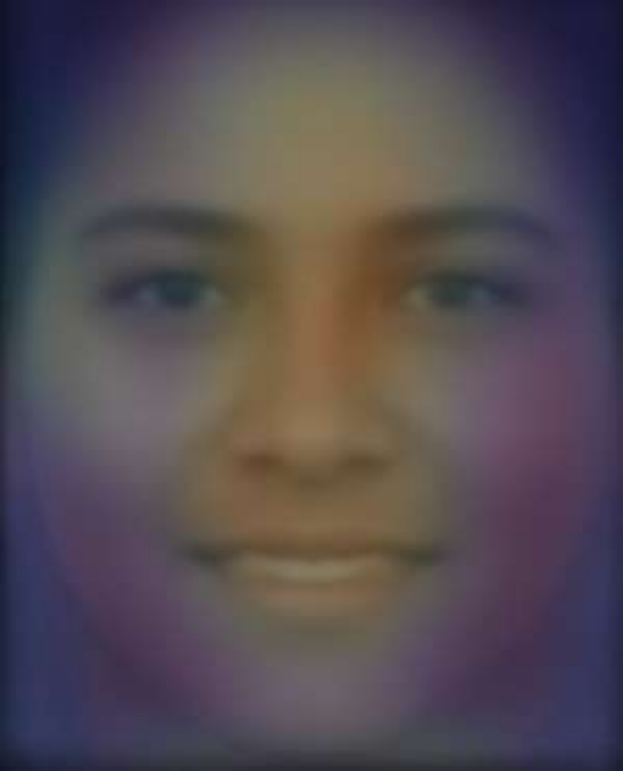}
		% \label{fig1:c}
	\end{subfigure}

	\begin{subfigure}{0.23\columnwidth}
		\includegraphics[width= \textwidth, height=2.5cm]{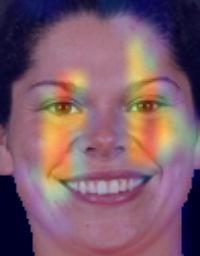}
		% \label{fig1:a}
	\end{subfigure}
     \begin{subfigure}{0.23\columnwidth}
		\centering
		\includegraphics[width= \textwidth, height=2.5cm]{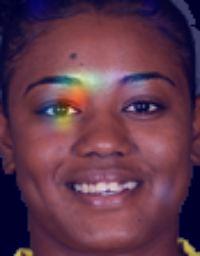}
		% \label{fig1:b}
	\end{subfigure}
	\begin{subfigure}{0.23\columnwidth}
		\centering
		\includegraphics[width= \textwidth, height=2.5cm]{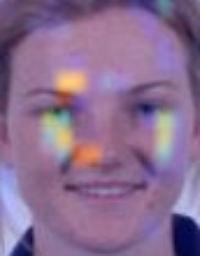}
		% \label{fig1:c}
	\end{subfigure}
    \begin{subfigure}{0.23\columnwidth}
		\centering
		\includegraphics[width= \textwidth, height=2.5cm]{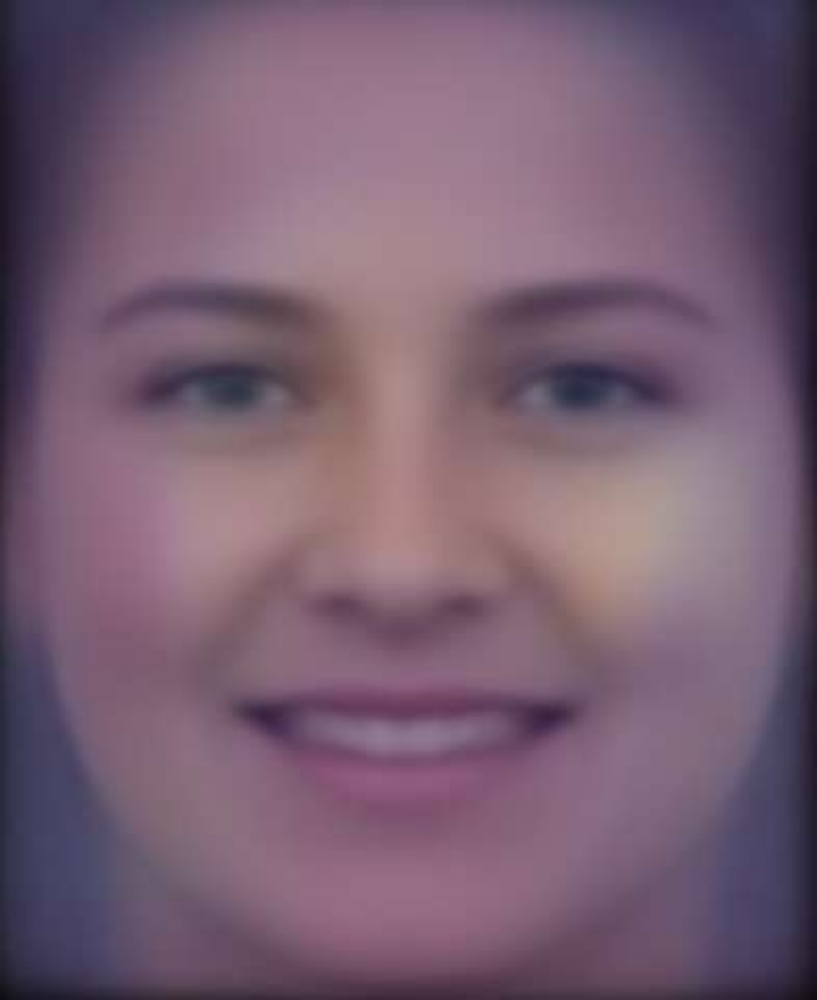}
		% \label{fig1:c}
	\end{subfigure}
    
	\begin{subfigure}{0.23\columnwidth}
		\includegraphics[width= \textwidth, height=2.5cm]{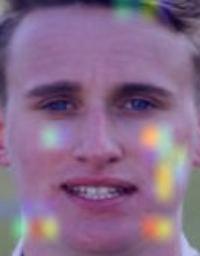}
		% \label{fig1:a}
	\end{subfigure}
     \begin{subfigure}{0.23\columnwidth}
		\centering
		\includegraphics[width= \textwidth, height=2.5cm]{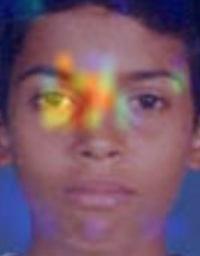}
		% \label{fig1:b}
	\end{subfigure}
	\begin{subfigure}{0.23\columnwidth}
		\centering
		\includegraphics[width= \textwidth, height=2.5cm]{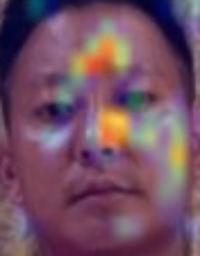}
		% \label{fig1:c}
	\end{subfigure}
    \begin{subfigure}{0.23\columnwidth}
		\centering
		\includegraphics[width= \textwidth, height=2.5cm]{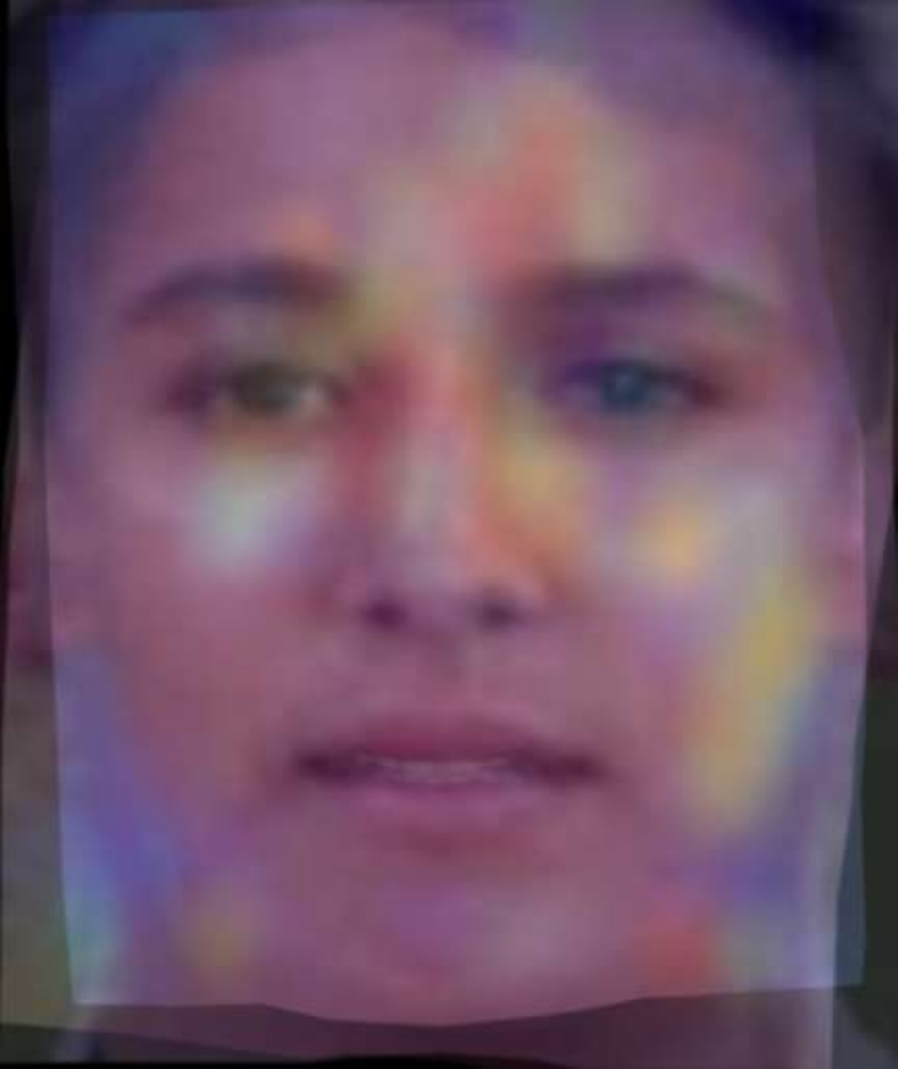}
		% \label{fig1:c}
	\end{subfigure}

	\caption{\footnotesize \textbf{Example activation maps from the Grad-CAM analysis of the ORIG set for Deepface. The images are ordered as -- Row 1: Males \textit{correctly predicted} as male (New Zealand, Pakistan, Bangladesh), Row 2: Females \textit{incorrectly predicted} as male (New Zealand, West Indies, South Africa), Row 3: Females \textit{correctly predicted} as female (England, West Indies, Australia) and, Row 4: Males \textit{incorrectly predicted} as female (England, India, India). For the images classified as male (first two rows), there is a more systematic region of interest, whereas the region of interest seems random for images classified as female (the last two rows). The last column in each row corresponds to the average Grad-CAM activation maps indicating the generalizability of our analysis.}}
	\label{fig:gradcam_example}
\end{figure}

Figure~\ref{fig:gradcam_example} shows the Grad-CAM activation maps for the Deepface predictions using 12 representative images from the \textsc{FARFace} dataset. We observe that the FRS model's attention heatmap when classifying images as male is very different compared to when it classifies them as female.
For the male classifications (the first two rows in Figure~\ref{fig:gradcam_example}), the model's focus is on a narrow vertical region stretching from the forehead to the mouth, with the most important region being the nose, as seen from the color grading on the activation map. On the other hand, when the model predicts an image as female, the region of interest varies and random pixels are highlighted in the image. The patterns are independent of the region to which the individual belongs (GN or GS). We present some representative images here, but the observations can be generalized over all images in the dataset (last column in Fig.~\ref{fig:gradcam_example}).

\noindent \textbf{Takeaways}:
\textit{There is a significant difference in the activation maps of the Deepface model's classification for the two genders. For the male classifications, it has a more systematic region of interest, whereas the region of interest is random for images classified as female.}

\subsection{Mitigating biases in FRSs (RQ3)}
\begin{table*}[!ht]
    \scriptsize
    \begin{center}        
    \begin{tabular}{|c|c||c|c||c|c||c|c||c|c||c|c||c|c||c|c|}
    \hline
    \multirow{3}{*}{\textbf{Type}} & \multirow{3}{*}{\textbf{Region}} & \multicolumn{2}{c||}{\multirow{2}{*}{\textbf{Pre-trained}}} & \multicolumn{6}{|c||}{\textbf{One-shot}}  & \multicolumn{6}{|c|}{\textbf{Two-shot}}\\ \cline{5-16}
        & & \multicolumn{2}{c||}{} & \multicolumn{2}{c||}{\textbf{0\% Adv}} & \multicolumn{2}{|c||}{\textbf{50\% Adv}} & \multicolumn{2}{|c||}{\textbf{100\% Adv}} & \multicolumn{2}{|c||}{\textbf{0\% Adv}} & \multicolumn{2}{|c||}{\textbf{50\% Adv}} & \multicolumn{2}{|c|}{\textbf{100\% Adv}}\\ \cline{3-16}
        ~ & ~ & M & F & M & F & M & F & M & F & M & F & M & F & M & F \\ \hline\hline
        \multirow{2}{*}{ORIG} & GN & \textbf{100} & 70 & 86.04 & \textbf{100} & 88.96 & \textbf{100} & 88.96 & \textbf{100} & 93.33 & \textbf{99.58} & 91.67 & \textbf{99.58} & 88.75 & \textbf{100} \\ \cline{2-16}
        ~ & GS & \textbf{100} & 30 & 91.25 & 88.75 & 94.38 & 86.25 & 94.58 & 86.25 & 96.67 & 80.42 & 95.21 & 82.08 & 92.71 & 86.67 \\ \hline\hline
        \multirow{2}{*}{RGB$_{0.3}$} & GN & \textbf{100} & 51.25 & 91.88 & 91.67 & 85.21 & \textbf{97.5} & 83.96 & \textbf{98.33} & 94.58 & 91.67 & 89.58 & \textbf{97.5} & 87.5 & \textbf{95.42} \\ \cline{2-16}
        ~ & GS & \textbf{100} & 16.25 & \textbf{95.42} & 65.83 & 91.46 & 77.92 & 87.29 & 79.17 & \textbf{98.54} & 59.17 & 96.25 & 72.5 & 91.04 & 75.42 \\ \hline
    \end{tabular}
    \caption{ \footnotesize \textbf{Average male and female gender prediction accuracies for the images belonging to the ORIG and the RGB$_{0.3}$ set from Global North (GN) and Global South (GS) on a test set of 480 images for few-shot fine-tuning on Deepface. One-shot and two-shot refer to the number of fine-tuning examples chosen from the  dataset for each gender from each country. 0\%, 50\% and 100\% refer to the number of RGB$_{0.3}$ adversarial examples used during fine-tuning, the rest being from the ORIG set. Accuracies for females improve significantly, irrespective of the region. One-shot performs better on the ORIG set and two-shot on the RGB$_{0.3}$ set. Max values for each set and mitigation setup are in bold}.}
    \label{tab:finetuning}
    \end{center}
\end{table*}

Previously, we presented the audit study for FRSs on the \textsc{FARFace} dataset. We observed that all FRSs report disparate performances for females, specifically from the Global South. We now study the results for the two bias mitigation strategies (discussed earlier) for the Deepface open-source FRS -- (1)~\textbf{few-shot learning}, and (2)~\textbf{contrastive learning}. 

\noindent \textbf{Results for few-shot learning}: In Table~\ref{tab:finetuning}, we report the results for the few-shot learning setup (more results are shared in the Supplementary section). 
These are compared against the baseline values from the pre-trained zero-shot setup on the held out test set of 480 images. In the pre-trained model, the accuracy of females is inferior to males, irrespective of the region on the ORIG as well as the RGB$_{0.3}$ set. When the FRS is fine-tuned with one-shot examples, the accuracy for females improves significantly -- by 30\% for Global North (giving \textit{100\% accuracy}) and by more than 55\% for Global South in the ORIG set. This improvement is similar for the RGB$_{0.3}$ test set ($> 40\%$). Finally, we see that the accuracy for females on the RGB set improves with an increase in adversarial examples, whereas the opposite is true for the ORIG set. We see similar trends for the two-shot scenario, wherein the absolute values are lower than the one-shot scenario.
Here, the accuracy on both sets improves with an increasing number of RGB$_{0.3}$ fine-tuning examples. 

Next, observing the accuracy for predicting the male gender for the one-shot learning scenario, we see that the accuracy decreases from 100\% in the pre-trained model by as much as 14\% (ORIG) -- 16\% (RGB$_{0.3}$). Increasing the ratio of adversarial examples in the fine-tuning set reduces the accuracy for male prediction, the opposite of what we observe for females. Next, in the two-shot scenario, the accuracy of predicting males increases again -- maximum of 98.54\% for males from the Global South in the RGB$_{0.3}$ set.

Finally, we note two consistent trends -- (i) the accuracy for gender prediction of females from the Global South is always lower than that of the females from the Global North and, (ii) the accuracy on females (independent of the fine-tuning setting) is always higher than the males except in the RGB$_{0.3}$ set, when fine-tuned using ORIG examples only. This shows that only few examples are needed to significantly reduce the bias and improve the accuracy of the marginalized class. 

\begin{table}[!t]
    \tiny
    \centering
    \begin{tabular}{|c|c|c|c|c|c|}
    \hline
        \textbf{Model} & \textbf{CelebSET} & \textbf{CFD-USA} & \textbf{CFD-India} & \textbf{CFD-MR} & \textbf{Fairface} \\ \hline
        AWS & 99.5\% & 97.32\% & 98.59\% & 97.73\% & 92.11\% \\ \hline
        FPP & 99.06\% & 93.47\% & 92.96\% & 90.91\% & 90.83\% \\ \hline
        MSFT & 99.31\% & 99.83\% & 100\% & 100\% & 80.01\% \\ \hline
        LIBFC & 85.94\% & 72.70\% & 79.58\% & 76.14\% & 76.67\% \\ \hline
        DPFC & 82.06\% & 79.56\% & 73.94\% & 67.05\% & 71.67\% \\ \hline \hline
        DPFC-FT & 97.75\% & 85.26\% & 93.66\% & 89.77\% & 84.62\% \\ \hline
    \end{tabular}
    \caption{ \footnotesize \textbf{Benchmark on other datasets for all FRSs and for 2-shot fine-tuned Deepface using the FARFace dataset (DPFC-FT). The Deepface model reports high accuracies on all datasets after fine-tuning proving generalizability of our technique.}}
    \label{tab:ft_other_datasets}
\end{table}

\begin{table}[!t]
    \centering
    \tiny
    \begin{tabular}{| c | c | c | c | c | c |}
    \hline
    \textbf{Model} & \textbf{CelebSET} & \textbf{CFD-USA} & \textbf{CFD-India} & \textbf{CFD-MR} & \textbf{Fairface} \\
    \hline
    AWS & 0.75\% & 3.20\% & 3.85\% & 3.23\% & 5.31\% \\
    \hline
    FPP & 0.87\% & 12.04\% & 19.23\% & 12.90\% & 6.57\% \\
    \hline
    MSFT & 0.13\% & 0.33\% & 0\% & 0\% & 6.83\% \\
    \hline
    LIBFC & 0.12\% & 33.65\% & 49.70\% & 22.96\% & 22.47\% \\
    \hline
    DPFC & 29.63\% & 39.07\% & {71.15\%} & 14.90\% & 56.09\% \\
    \hline \hline
    DPFC-FT & 0.74\% & 10.56\% & 8.21\% & 12.78\% & 12.28\% \\
    \hline
    \end{tabular}
    \caption{\footnotesize \textbf{Absolute disparity between Males \& Females on other datasets for all FRSs and for 2-shot fine-tuned Deepface using the FARFace dataset (DPFC-FT). The Deepface model after fine-tuning reports significantly lower disparity on all datasets.}}
    \label{tab:ft_other_datasets_disp}
    \end{table}

\noindent \textbf{Results for generalizability to other datasets}: We benchmark all the five pre-trained FRSs and the Deepface model fine-tuned on 2-shot images from the FARFace dataset on five highly popular diverse face datasets viz. CelebSET~\cite{raji2020saving}, CFD-USA~\cite{ma2015chicago}, CFD-India~\cite{lakshmi2021india}, CFD-MR~\cite{ma2020chicago} and Fairface~\cite{karkkainenfairface}. From Table~\ref{tab:ft_other_datasets} we see that while the three commercial systems report significantly high accuracies on all datasets, Libfaceid and Deepface lag far behind. Thus, these open-source models, despite being trained on faces from the Global North do not generalize well to even these datasets. 
Our fine-tuned model (DPFC-FT), on the other hand, reports a significantly high accuracy for other datasets as well exhibiting good generalizability prowess. We also observe a reduction in disparity in gender accuracy for all datasets for our fine-tuned model as compared to the pre-trained Deepface model, in Table~\ref{tab:ft_other_datasets_disp}. For all datasets except CFD-MR, the disparity reduction ranges from 29-60\%. Thus, we can state that \textit{models fine-tuned on images from the Global South can be re-used in other regions without loss in performance.}

\begin{figure}[!t]
	\centering
	\begin{subfigure}{0.23\columnwidth}
		\includegraphics[width= \textwidth, height=2.5cm]{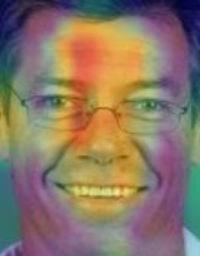}
		% \label{fig1:a}
	\end{subfigure}
	\begin{subfigure}{0.23\columnwidth}
		\includegraphics[width= \textwidth, height=2.5cm]{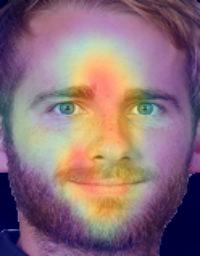}
		% \label{fig1:a}
	\end{subfigure}
     \begin{subfigure}{0.23\columnwidth}
		\centering
		\includegraphics[width= \textwidth, height=2.5cm]{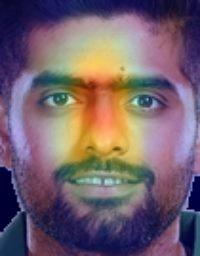}
		% \label{fig1:b}
	\end{subfigure}
	\begin{subfigure}{0.23\columnwidth}
		\centering
		\includegraphics[width= \textwidth, height=2.5cm]{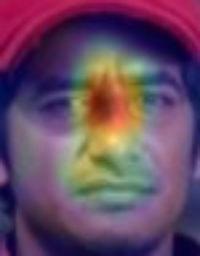}
		% \label{fig1:c}
	\end{subfigure}

    \begin{subfigure}{0.23\columnwidth}
		\includegraphics[width= \textwidth, height=2.5cm]{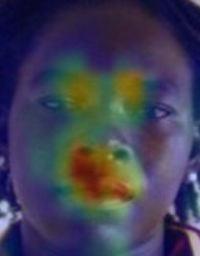}
		% \label{fig1:a}
	\end{subfigure}
	\begin{subfigure}{0.23\columnwidth}
		\includegraphics[width= \textwidth, height=2.5cm]{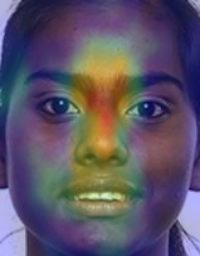}
		% \label{fig1:a}
	\end{subfigure}
     \begin{subfigure}{0.23\columnwidth}
		\centering
		\includegraphics[width= \textwidth, height=2.5cm]{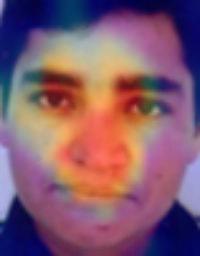}
		% \label{fig1:b}
	\end{subfigure}
	\begin{subfigure}{0.23\columnwidth}
		\centering
		\includegraphics[width= \textwidth, height=2.5cm]{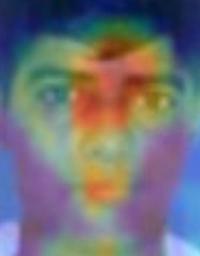}
		% \label{fig1:c}
	\end{subfigure}

     \begin{subfigure}{0.23\columnwidth}
		\includegraphics[width= \textwidth, height=2.5cm]{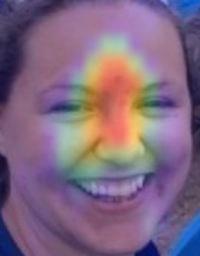}
		% \label{fig1:a}
	\end{subfigure}
	\begin{subfigure}{0.23\columnwidth}
		\includegraphics[width= \textwidth, height=2.5cm]{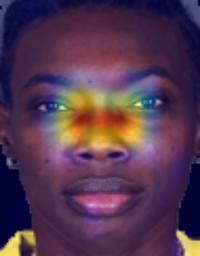}
		% \label{fig1:a}
	\end{subfigure}
     \begin{subfigure}{0.23\columnwidth}
		\centering
		\includegraphics[width= \textwidth, height=2.5cm]{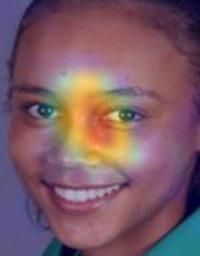}
		% \label{fig1:b}
	\end{subfigure}
	\begin{subfigure}{0.23\columnwidth}
		\centering
		\includegraphics[width= \textwidth, height=2.5cm]{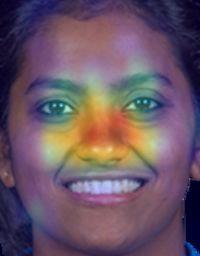}
		% \label{fig1:c}
	\end{subfigure}
    
    \begin{subfigure}{0.23\columnwidth}
		\includegraphics[width= \textwidth, height=2.5cm]{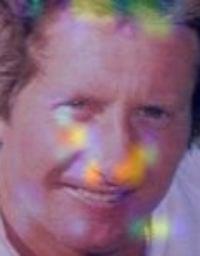}
		% \label{fig1:a}
	\end{subfigure}
	\begin{subfigure}{0.23\columnwidth}
		\includegraphics[width= \textwidth, height=2.5cm]{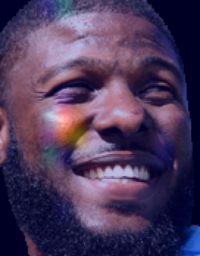}
		% \label{fig1:a}
	\end{subfigure}
     \begin{subfigure}{0.23\columnwidth}
		\centering
		\includegraphics[width= \textwidth, height=2.5cm]{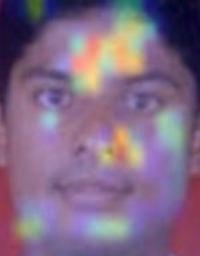}
		% \label{fig1:b}
	\end{subfigure}
	\begin{subfigure}{0.23\columnwidth}
		\centering
		\includegraphics[width= \textwidth, height=2.5cm]{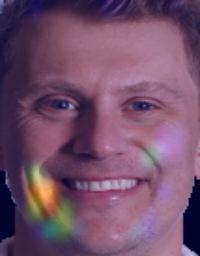}
		% \label{fig1:c}
	\end{subfigure}

    \begin{subfigure}{0.23\columnwidth}
		\includegraphics[width= \textwidth, height=2.5cm]{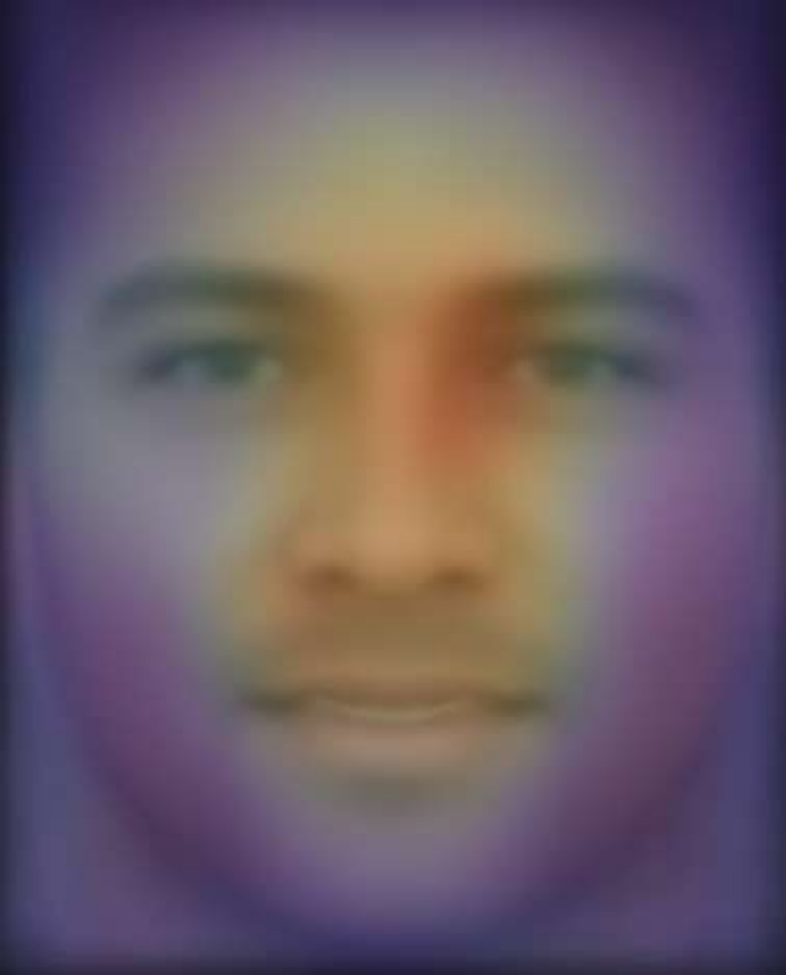}
		% \label{fig1:a}
	\end{subfigure}
	\begin{subfigure}{0.23\columnwidth}
		\includegraphics[width= \textwidth, height=2.5cm]{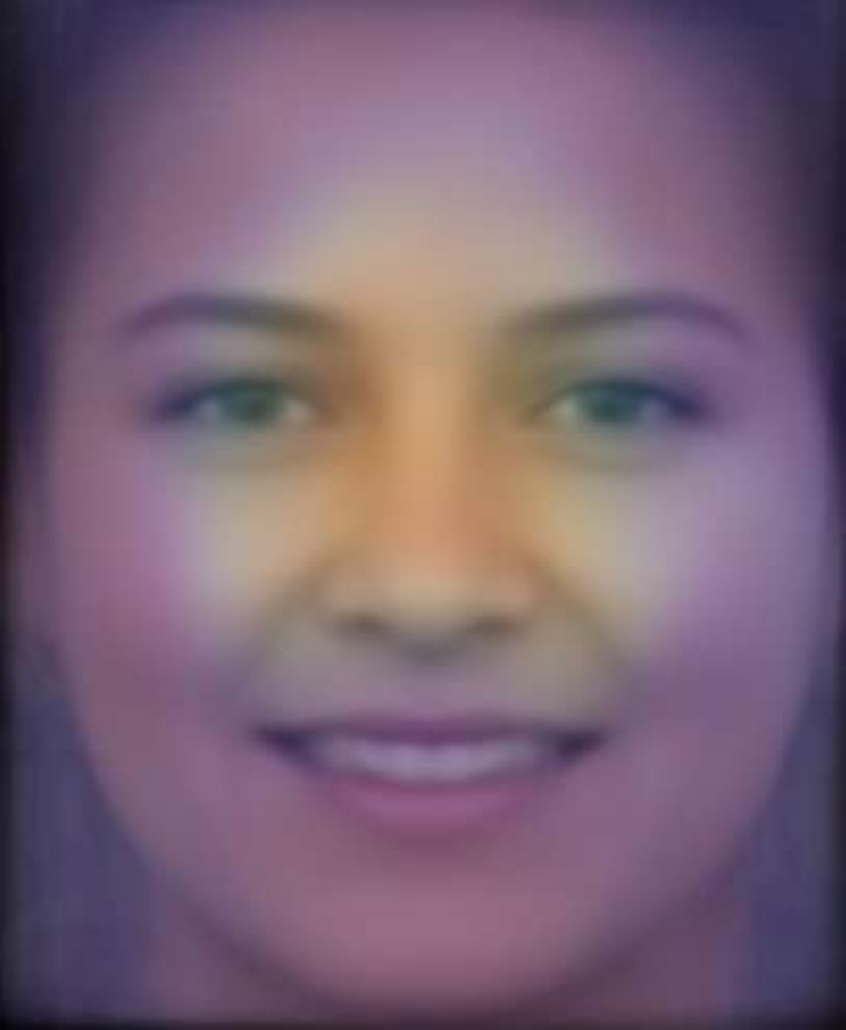}
		% \label{fig1:a}
	\end{subfigure}
     \begin{subfigure}{0.23\columnwidth}
		\centering
		\includegraphics[width= \textwidth, height=2.5cm]{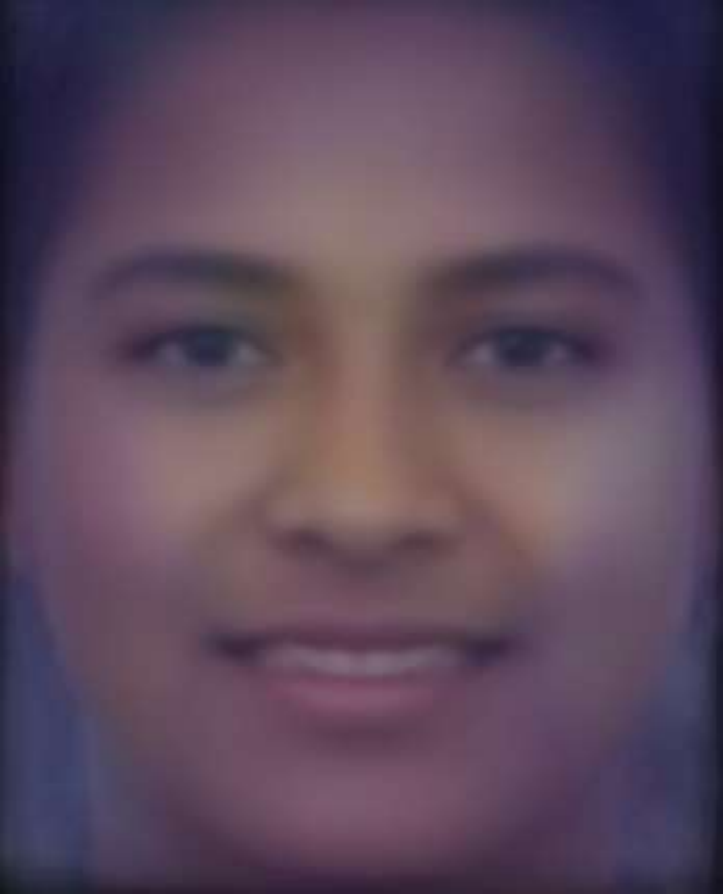}
		% \label{fig1:b}
	\end{subfigure}
    \begin{subfigure}{0.23\columnwidth}
		\centering
		\includegraphics[width= \textwidth, height=2.5cm]{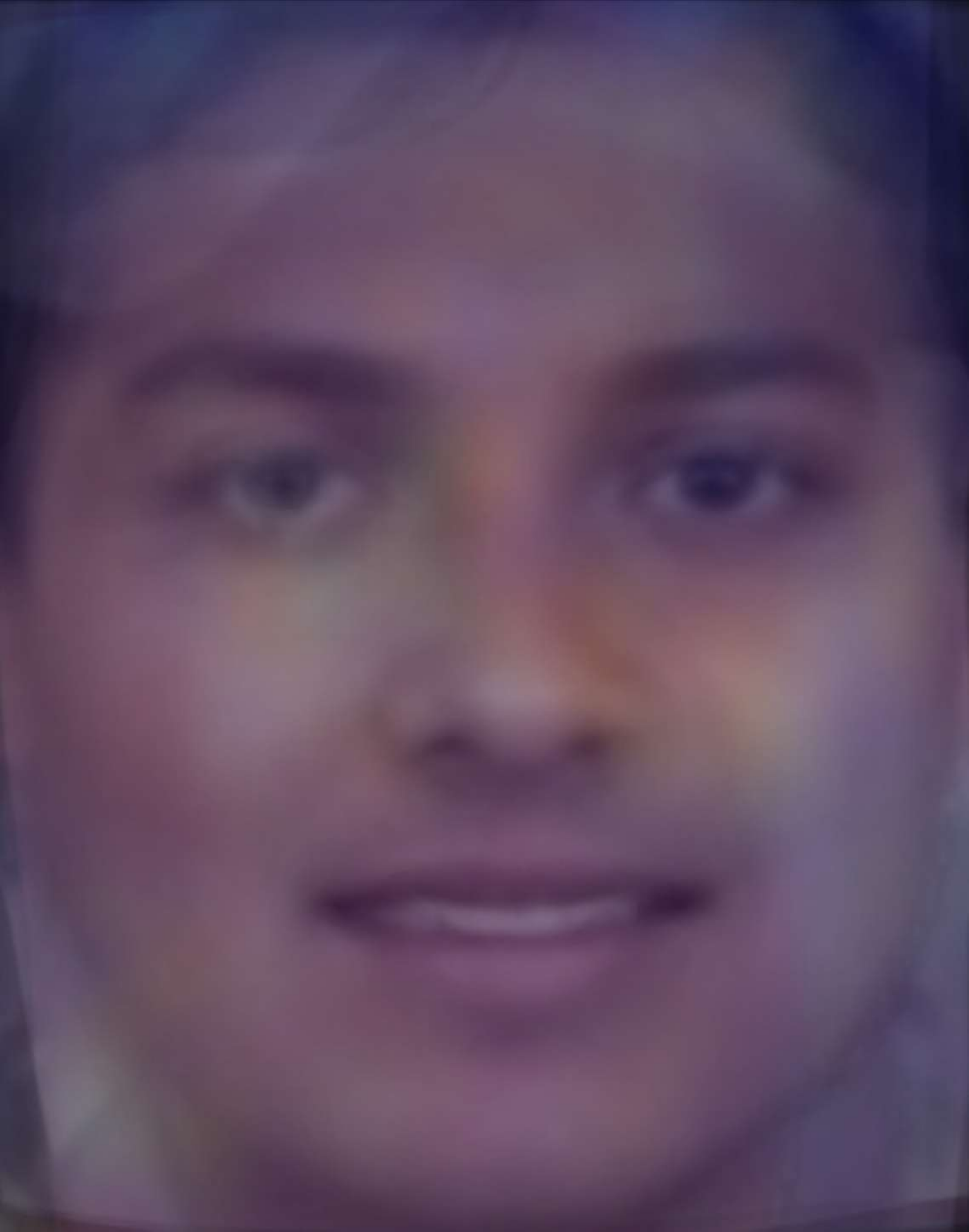}
		% \label{fig1:b}
	\end{subfigure}
	\caption{\footnotesize \textbf{Example activation maps from the Grad-CAM analysis of the ORIG set (held-out test set) for Deepface after two-shot fine-tuning on the ORIG set. The images are ordered as -- Row 1: Males \textit{correctly predicted} as male (Australia, New Zealand, Pakistan, Bangladesh), Row 2: Females \textit{incorrectly predicted} as male (West Indies, Bangladesh, Pakistan, India), Row 3: Females \textit{correctly predicted} as female (England, West Indies, South Africa, India) and, Row 4: Males \textit{incorrectly predicted} as female (Australia, West Indies, India, England). Row 5 has the average Grad-CAM activation maps for the images of males correctly predicted as male, females correctly predicted as females, females incorrectly predicted as males, males incorrectly predicted as females; it is apparent that there is a more systematic focus on the nose when females are being correctly predicted now.}}
	\label{fig:gradcam__after_ft_example}
\end{figure}

\noindent\noindent\textbf{Grad-CAM analysis}: We evaluate the Grad-CAM activation maps of the held-out test set for the following setup: two-shot fine-tuning with all training examples chosen from the ORIG set. Here we observe a change in the highlighted area of the image when the model predicts the female gender -- the model's region of interest is around the \textit{nose} of the person, as opposed to random areas within the image. This indicates a clear change in the model's decision making process, specially for the true positive data points. Example activation maps are shown in Figure~\ref{fig:gradcam__after_ft_example}. Consider activation map in row 2, col 1 of Fig.~\ref{fig:gradcam_example} and in row 3, col 2 of Fig.~\ref{fig:gradcam__after_ft_example}-- both belong to the same female and there is a clear change in the area of interest. This analysis generalizes over the entire held-out test set as is evident from the last row of Fig.~\ref{fig:gradcam__after_ft_example}. 

\noindent \textbf{Results for contrastive learning}: Finally, we also experiment with a \textit{novel} contrastive learning setup. Recall, for the computation of the triplet loss for an anchor point (any data point in a given class) ($x_a$), the positive example ($x_{a}^{+}$) is an adversarial image of the same individual from RGB$_{0.3}$, whereas the negative example ($x_{a}^{-}$) is an image of another person from the ORIG set of the opposite gender (more results for other variants are shared in the Supplementary section).
Table~\ref{tab:contrastive} shows the accuracy in the ORIG set for gender prediction increases significantly for females from both Global North (18.75\%) and Global South (50\%). The increase is more significant for images in the RGB$_{0.3}$ set with an increment of 37.5\% in the Global North and 60\% in the Global South, thereby justifying our choice of adversarial samples in the learning setup. In compensation, the accuracy for males drops, with a max drop of $\approx 20\%$. We do not observe a large reduction in bias between the different groups here.

\begin{table}[!t]
\footnotesize
    \begin{center}        
    \begin{tabular}{|c|c||c|c||c|c|}
    \hline
    \multirow{2}{*}{\textbf{Type}} & \multirow{2}{*}{\textbf{Region}} & \multicolumn{2}{|c||}{\textbf{Pre-trained}} & \multicolumn{2}{|c|}{\textbf{Contrastive}} \\
    \cline{3-6}
     & & M & F & M & F \\ \hline \hline
    \multirow{2}{*}{ORIG} & GN & \textbf{100} & 70 & 82.5 & 88.75 \\ \cline{2-6}
    & GS & \textbf{100} & 30 & \textbf{90} & 80 \\ \hline\hline
    \multirow{2}{*}{RGB$_{0.3}$} & GN & \textbf{100} & 51.25 & 81.88 & \textbf{88.75} \\ \cline{2-6}
    & GS & \textbf{100} & 16.25 & 87.5 & 76.25 \\ \hline
    \end{tabular}
    \caption{\footnotesize \textbf{Average male and female gender prediction accuracies for the contrastive learning setup on test sets of 480 images. Female accuracies from both regions improve significantly. Accuracy for females from Global North is higher and for males is lower for both test sets}.}
    \label{tab:contrastive}
    \end{center}
\end{table}

\noindent \textbf{Fairness vs. accuracy}: As this is a binary classification task, we hypothesize that the drop in accuracy for males and significant improvement in accuracy for females is a result of an angular shift in the separating hyperplane, causing a small number of previously classified males to now be classified as female, but a larger number of females to be classified correctly. Thus we see that the cost of fairness (reduction in disparity between the performance for the two genders) is a drop in accuracy for males, the majority class, which is typically a standard observation in the literature. A more in-depth mathematical study to bound this shift in accuracy, while maintaining fairness is deferred to a future study.

\noindent \textbf{Takeaways:}
\textit{(a) The model is highly receptive to few-shot fine-tuning setups showing a significant improvement in accuracy, especially for females in the Global South,} and, \textit{(b) Adversarial examples improve the accuracy as well as the robustness of the fine-tuning setup.}

\subsection{Results of red-teaming task (country prediction)}
\begin{table}[!t]
    \tiny
    \centering
    \begin{tabular}{|c|c|c|c|c|c|c|c|}
    \hline
    \multirow{2}{*}{\textbf{Scheme}} & \multirow{2}{*}{\textbf{Reg.}} & \multicolumn{6}{|c|}{\textbf{Type}} \\
    \cline{3-8}
        &  & \textbf{ORIG} & \textbf{RGB}$_{0.3}$ & \textbf{RGB}$_{0.5}$ & \textbf{GRAY} & \textbf{SPRD} & \textbf{MASK} \\ \hline\hline
      \multirow{2}{*}{\textbf{F}}  & GN & 54.17 & 60.83 & 64.17 & 53.33 & 52.50 & 62.50 \\ \cline{2-8}
        & GS & 75 & 53.33 & 41.67 & 76.11 & 49.44 & 57.22 \\ \hline\hline
       \multirow{2}{*}{\textbf{C}} & GN & 87.92 & 83.33 & 86.25 & 86.25 & 81.25 & 72.38 \\ \cline{2-8}
        & GS & 90.56 & 89.44 & 86.67 & 80.56 & 92.22 & 60.42 \\ \hline        
    \end{tabular}
    \caption{\footnotesize \textbf{Average country prediction accuracies for the two regions, for the fine-tuning (F) and contrastive learning (C) setup. In the fine-tuning setup, first, the model is fine-tuned on Geofaces, followed by a 2-shot fine-tuning on FARFace. In the second setup, the model is trained using contrastive learning first on the Geofaces dataset and followed by a second round of contrastive learning on the FARFace dataset. The two rounds of contrastive learning produce the best results.}}
    \label{tab:ft_contrastive_country}
\end{table}

The results for predicting the country from face image are shown in Table~\ref{tab:ft_contrastive_country}. Here, we see that depending on the adversarial variant, one region reports a higher accuracy over the other. Overall, fine-tuning results are poor with accuracy values around 50\% only. The  2-stage contrastive learning setup produces the best and least disparate results. However, as a note of caution we would like to state that the premise of predicting one's country or ethnicity from their face image is a flawed one and has primarily been used for surveillance~\cite{mozur2019uyghur}. As an example, one can consider a country like India with its large ethnicity and individuals being profiled as belonging to other countries because of their facial phenotype, leading to unfair and biased outcomes.

\section{Concluding discussion}
\label{sec:discussions}

\noindent \textbf{Benchmark audit of FRSs}:
In our benchmark audit, we see that all models have a good accuracy on the ORIG set, but commercial models are less robust to adversarial inputs; all FRSs report disparate results against Global South females (generally dark skin tone). Such biases were observed previously on commercial FRSs~\cite{buolamwini2018gender,raji2020saving,jaiswal2022two}. Thus we observe both \textit{temporal} and \textit{emergent} biases against dark-skinned women~\cite{mehrabi2021survey} through our study. Our current study supports existing observations and necessitates the \textit{\textbf{need for continual temporal audits}} of FRSs, especially for adversarial real-world inputs. 

\noindent\textbf{Explanation of FRS outputs}:
Grad-CAM analysis on the Deepface FRS reinforced the audit observations that FRS models are better equipped to predict gender for males (the IMDB-Wiki~\cite{rothe2018deep} dataset has majority male faces~\cite{raji2020saving}); these observations generalize across countries. This leads us to the final part of our pipeline -- bias mitigation.

\noindent\textbf{Bias mitigation on Deepface}:
Our few-shot and contrastive learning approaches result in improved accuracy for females in both regions. Accuracy for the RGB$_{0.3}$ set also improves, resulting in increased model robustness. Few-shot fine-tuning performs better than contrastive learning overall, but we defer further exploration of the reasoning to a future study. Simple techniques like fine-tuning and contrastive learning improve accuracy while reducing bias, and adversarial examples improve robustness.

\noindent \textbf{Limitations and scope for future work}: 
While the current dataset has images from both Global South and Global North, the age related information and gender diversity (as the sport of cricket is male dominated) is missing. We would like to improve the dataset on these dimensions in the future. The current study focuses on classification tasks. We intend to extend it to retrieval tasks like face verification. We also plan to study the accuracy vs fairness trade-offs for our mitigation strategies and introduce better algorithms that reconciles between the two notions better.

\section*{Acknowledgements}
This research is supported partly by a European Research Council (ERC) Advanced Grant for the project ``Foundations for Fair Social Computing", funded under the European Union's Horizon 2020 Framework Programme (grant agreement no. 789373), and by a grant from the Max Planck Society through a Max Planck Partner Group at IIT Kharagpur. S. Jaiswal is supported by the Prime Minister's Research Fellowship, Govt. of India.

\section*{Ethics statement}
In this work, we have introduced a new large-scale face dataset, curated from images of cricket players. We would like to state here that we use the gender labels already publicly available on the \texttt{ESPNCricInfo} website and do not attempt to modify these or any other ground truth attributes. We do not collect any private information. Our study is limited to binary gender classification, but this is primarily due to only binary gender labels -- male \& female being available in all face datasets (ours and in literature). We do not intend to propagate the perspective that gender is a binary label; we acknowledge that gender is a fluid spectrum, and hence all gender labels in our study are to be interpreted as \textit{perceived} gender. 

Next, we note that our dataset is unbalanced, with a larger ratio of males than females. This is no surprise as it is reflective of gender imbalance observed in many professions, including sports~\cite{toi_puru_2023} and is a societal issue that needs deeper reflection and redressal through governmental intervention. Multiple prior open-source models have also been trained on such imbalanced datasets as is evident from their bias against females in audit studies. We do not condone this and, as such, to prevent further misuse, will only share a balanced subset of our dataset for any and all research purposes. 
We also audit on a balanced subset of images and identify that similar biases as shown here persist (Results are in an extended version of the paper\footnotemark[6]). Even in our bias mitigation experiments, we have only used balanced subsets (for both fine-tuning and contrastive learning). 

Images from the Global South can include additional sensitive features, for example, caste or religion in South Asia, which are correlated to one's socio-economic position in these societies. These attributes intersect with sensitive features like skin tone and can lead to newer skews beyond Global North and South. We acknowledge this as a limitation of our dataset and recommend future users of the dataset to be mindful when performing sociotechnical analyses.

Finally, our red teaming experiment has been done to show the relative ease of misusing FRSs for controversial and morally ambiguous experiments like predicting the country of a person from their face. Ironically, the results from this experiment too are biased, thus showing how models trained on datasets from the Global North encode strong biases. We will not be releasing either the code or the specific subset of images used for this experiment.

Our primary motivation for this work stems from the under-representation of the Global South in the larger AI development landscape and over-representation in the AI testbed landscape. We hope our dataset and bias mitigation algorithms will break this hegemony and provide some balance.

\bibliography{main}

\clearpage

\section*{Supplementary Material}
\renewcommand{\thesubsection}{\Alph{subsection}}
Here we present additional results for all experiments that we conducted and presented as part of our main submission. The overall accuracy and disparity for the benchmark audit, and accuracies for multiple combinations for few-shot \& contrastive learning setups are shown in order as follows.

\subsection{Benchmark audit of FRSs}

\begin{table*}[!ht]
    \scriptsize
    \begin{center}
    \begin{tabular}{|c|c||c|c||c|c||c|c||c|c||c|c||c|c|}
    \hline
        \multirow{3}{*}{\textbf{FRS}} & \multirow{3}{*}{\textbf{Region}} & \multicolumn{12}{c|}{\textbf{Image Type}} \\ \cline{3-14}
        &  & \multicolumn{2}{c||}{\textbf{ORIG}} & \multicolumn{2}{c||}{\textbf{GREY}} & \multicolumn{2}{c||}{\textbf{RGB$_{0.3}$}} & \multicolumn{2}{c||}{\textbf{RGB$_{0.5}$}} & \multicolumn{2}{c||}{\textbf{SPRD}} & \multicolumn{2}{c|}{\textbf{MASK}} \\ \cline{3-14}
        ~ & ~ & \textbf{M} & \textbf{F} & \textbf{M} & \textbf{F} & \textbf{M} & \textbf{F} & \textbf{M} & \textbf{F} & \textbf{M} & \textbf{F} & \textbf{M} & \textbf{F} \\ \hline\hline
        \multirow{2}{*}{\textbf{AWS}} &  GN & 99.04 & 95.37 & 99.6 & 93.95 & 93.68 & 88.61 & 77.66 & 79.18 & 97.88 & 92.88 & 95.67 & 78.29 \\ \cline{2-14}
        ~ &  GS & 98.51 & 82.11 & 98.17 & 85.91 & 96.6 & 72.36 & 77.65 & 60.98 & 98.92 & 84.28 & 95.44 & 63.41 \\ \hline\hline
        \multirow{2}{*}{\textbf{MSFT}} &  GN & 98.92 & 96.26 & 98.81 & 96.26 & 94.14 & 91.64 & 68.49 & 71.71 & 97.84 & 92.17 & 48.12 & 43.42 \\ \cline{2-14}
        ~ &  GS & 99 & 88.35 & 99.11 & 85.64 & 91.95 & 73.44 & 69.97 & 58.54 & 98.3 & 82.38 & 52.32 & 50.95 \\ \hline\hline
        \multirow{2}{*}{\textbf{FPP}} &  GN & 99.64 & 89.68 & 99.7 & 84.16 & 59.24 & 85.05 & 23.6 & 53.91 & 92.11 & 87.72 & 86.56 & 75.27 \\ \cline{2-14}
        ~ &  GS & 99.74 & 61.25 & 99.74 & 53.93 & 77.8 & 69.11 & 45.59 & 48.51 & 97.35 & 67.21 & 86.57 & 58.54 \\ \hline\hline
        \multirow{2}{*}{\textbf{DPFC}} &  GN & 99.9 & 75.98 & 99.95 & 78.83 & 100 & 47.51 & 100 & 15.48 & 99.8 & 54.63 & 96.41 & 46.62 \\ \cline{2-14}
        ~ &  GS & 99.96 & 26.29 & 99.96 & 29.27 & 100 & 12.74 & 99.96 & 2.98 & 99.89 & 17.62 & 98 & 18.16 \\ \hline\hline
         \multirow{2}{*}{\textbf{LIBFC}} &  GN & 96.88 & 30.25 & 94.24 & 22.42 & 84.64 & 65.48 & 77.72 & 68.86 & 94.13 & 38.26 & 95.77 & 5.52 \\ \cline{2-14}
        ~ &  GS & 98.55 & 15.18 & 95.88 & 15.99 & 91.2 & 42.82 & 86.93 & 43.36 & 97.11 & 25.75 & 96.77 & 5.69 \\ \hline
    \end{tabular}
    \caption{\bf Overall accuracy results for all FRSs with an equal number of samples (931 males and females) for both males and females (corresponds to Table~\ref{tab:regionwisegenderacc} in the paper). We take 5 random samples of 931 males and the fixed 931 females and present the average accuracies here.}
    \label{tab:overall_balanced_accuracy}
    \end{center}
\end{table*}

In Table~\ref{tab:overall_balanced_accuracy} we present the overall accuracy on the task of gender prediction for all FRSs on all image sets, in both regions, for a balanced set of Males and Females (931 each). From the table we see that all FRSs report higher accuracies for males as compared to females in both the regions. As these are balanced sets, this reconfirms our observation about the bias in these FRSs. The lowest accuracies for females are reported for the Masked adversarial input, with disparity in accuracy being more than 90\%. On average, the accuracy for males in the Global South is higher than in the Global North. 

\begin{table}[!ht]
    \scriptsize
    \begin{center}
    \begin{tabular}{|c|c|c|c|c|c|}
    \hline
        \multirow{2}{*}{\textbf{Image Type}} & \multicolumn{5}{|c|}{\textbf{FRS Platform}}\\
        \cline{2-6}
        & \textbf{AWS} & \textbf{MSFT} & \textbf{FPP} & \textbf{DPFC} & \textbf{LIBFC} \\ \hline\hline
        \textbf{ORIG} & 8.65 & 5.93 & 21.2 & 43.56 & 73.58 \\ \hline
        \textbf{RGB$_{0.3}$} & 13.2 & 9.06 & -8.14 & 66.29 & 31.39 \\ \hline
        \textbf{RGB$_{0.5}$} & 5.83 & 2.26 & -16.22 & 89.46 & 23.58 \\ \hline
        \textbf{GREY} & 7.6 & 6.95 & 27.32 & 40.68 & 75.58 \\ \hline
        \textbf{SPRD} & 8.87 & 9.96 & 16.04 & 59.92 & 62.62 \\ \hline
        \textbf{MASK} & 22.88 & 4.72 & 17.86 & 61.89 & 91.48 \\ \hline
    \end{tabular}
    \caption{\bf Disparity in accuracy between males and females for the task of gender prediction for all FRSs, with an equal number of samples (931 males and females). The highest disparity is reported for Face++ commercial FRS and Libfaceid open-source FRS. The values are calculated by subtracting accuracy for females from the accuracy for males.}
    \label{tab:overall_balanced_mf_disparity}
    \end{center}
\end{table}

Finally, in Table~\ref{tab:overall_balanced_mf_disparity}, we look at the disparity between the prediction accuracy for males and females across all FRSs on all the image types. We immediately see that the open-source FRSs have a higher disparity (ranging from 24.85\% to 93.45\%) than the commercial ones (ranging from 4.22\% to 42.95\%). On observing the rows we see that high disparities are reported for all image types, with the highest differences reported for the RGB$_{0.5}$ and masked variant (highest disparities for two FRSs each). This table indicates the extent and consistency of disparity between males and females.

\subsection{Mitigating biases in FRSs}
In Tables~\ref{tab:ft_orig}--\ref{tab:ft_all_adv} we present the results for the all the different fine-tuning setups that we try as part of this study. In Table~\ref{tab:contrastive_supp} we also present the results for the two different setups of contrastive learning that we experiment with.

\noindent \textbf{Results for few-shot learning}: We experiment with a few-shot learning based fine-tuning to mitigate the biases observed in the \textsc{FARFace} dataset for the Deepface open-source FRS. Here we show the results on all image sets for one-shot and two-shot fine-tuning with images from the ORIG set (Table~\ref{tab:ft_orig}), RGB$_{0.3}$ set (Table~\ref{tab:ft_rgb03}), RGB$_{0.5}$ set (Table~\ref{tab:ft_rgb05}), greyscale set (Table~\ref{tab:ft_grey}), spread set(Table~\ref{tab:ft_sprd}), masked set (Table~\ref{tab:ft_mask}). 

We also create a fine-tuning setting where the one-/two-shot data points are chosen from a combination of ORIG set and all other adversarial variants. Recall from the main draft that for one-shot learning, one image is chosen from each gender for each country, thus giving us 16 images (two genders for 8 countries). Similarly 32 images for two-shot learning. Thus for one-shot fine-tuning, 8 data points (50\%) are chosen from the ORIG set and the other 8 data points (remaining 50\%) are chosen as follows - two from each of RGB$_{0.3}$, Greyscale, Spread and Masked. A similar combination is used for the two-shot fine-tuning as well. The results for this setting are presented in Table~\ref{tab:ft_all_adv}.

\begin{table}[!t]
    \tiny
    %\small
    \begin{center}
    \begin{tabular}{|c|c||c|c||c|c||c|c||}
    \hline
    \multirow{2}{*}{\textbf{Type}} & \multirow{2}{*}{\textbf{Region}} & \multicolumn{2}{|c||}{\textbf{Pre-trained}} & \multicolumn{2}{|c||}{\textbf{1-shot}} & \multicolumn{2}{|c||}{\textbf{2-shot}} \\
     \cline{3-8}
     & & M & F & M & F & M & F \\ \hline \hline
        \multirow{2}{*}{ORIG} & GN & 100 & 70 & 86.04 & \textbf{100} & 93.33 & \textbf{99.58} \\ \cline{2-8}
        ~ & GS & 100 & 30 & 91.25 & 88.75 & 96.67 & 80.42 \\ \hline\hline
        \multirow{2}{*}{RGB$_{0.3}$} & GN & 100 & 51.25 & 91.88 & 91.67 & 94.58 & 91.67 \\ \cline{2-8}
        ~ & GS & 100 & 16.25 & \textbf{95.42} & 65.83 & \textbf{98.54} & 59.17 \\ \hline\hline
        \multirow{2}{*}{RGB$_{0.5}$} & GN & 100 & 16.25 & 95 & 66.25 & 98.33 & 64.58 \\ \cline{2-8}
        ~ & GS & 99.38 & 2.5 & \textbf{98.75} & 30.83 & \textbf{98.54} & 32.92 \\ \hline\hline
        \multirow{2}{*}{GREY} & GN & 100 & 75 & 85.42 & \textbf{100} & 93.33 & \textbf{99.17} \\ \cline{2-8}
        ~ & GS & 100 & 30 & 92.08 & 92.92 & 97.29 & 86.67 \\ \hline\hline
        \multirow{2}{*}{SPRD} & GN & 99.38 & 57.5 & 87.92 & \textbf{99.58} & 92.71 & \textbf{97.08} \\ \cline{2-8}
        ~ & GS & 100 & 21.25 & 91.25 & 84.58 & 95 & 76.25 \\ \hline\hline
        \multirow{2}{*}{MASK} & GN & 99.38 & 42.5 & 88.96 & 87.5 & 95 & 86.67 \\ \cline{2-8}
        ~ & GS & 100 & 28.75 & \textbf{91.04} & 75 & \textbf{96.67} & 69.17 \\ \hline
    \end{tabular}
     \caption{\footnotesize \textbf{Average male and female gender prediction accuracies  on test sets of 480 images of all image types, for fine-tuning on images from the ORIG set. Max values for a given combination of image type and learning setting are in bold.}}
    \label{tab:ft_orig}
    \end{center}
\end{table}

In Table~\ref{tab:ft_orig}, we look at the results for fine-tuning with images from the ORIG set only. Here we see that while females report higher accuracies for the one-shot learning, males report higher accuracies for two-shot learning (observations for both genders are independent of the region under consideration). We also note that the highest accuracies reported by females are for those from the Global North on the ORIG, GREY and SPRD sets, whereas the highest accuracies reported by males are for those from the Global South on the two RGB sets and the MASK set. These observations hold true for both one-shot and two-shot learning. Finally we see that the accuracy for females always increases with the few-shot learning with the maximum increase being 63.33\% for females in the Global South in the SPRD set. 

\begin{table}[!t]
    \tiny
    \begin{center}
    \begin{tabular}{|c|c||c|c||c|c||c|c||}
    \hline
    \multirow{2}{*}{\textbf{Type}} & \multirow{2}{*}{\textbf{Region}} & \multicolumn{2}{|c||}{\textbf{Pre-trained}} & \multicolumn{2}{|c||}{\textbf{1-shot}} & \multicolumn{2}{|c||}{\textbf{2-shot}} \\
     \cline{3-8}
     & & M & F & M & F & M & F \\ \hline \hline
        \multirow{2}{*}{ORIG} & GN & 100 & 70 & 88.96 & \textbf{100} & 88.75 & \textbf{100} \\ \cline{2-8}
        ~ & GS & 100 & 30 & 94.58 & 86.25 & 92.71 & 86.67 \\ \hline\hline
        \multirow{2}{*}{RGB$_{0.3}$} & GN & 100 & 51.25 & 83.96 & \textbf{98.33} & 87.5 & \textbf{95.42} \\ \cline{2-8}
        ~ & GS & 100 & 16.25 & 87.29 & 79.17 & 91.04 & 75.42 \\ \hline\hline
        \multirow{2}{*}{RGB$_{0.5}$} & GN & 100 & 16.25 & 83.75 & \textbf{86.67} & 90.42 & 81.25 \\ \cline{2-8}
        ~ & GS & 99.38 & 2.5 & 85 & 60.83 & \textbf{91.04} & 55.42 \\ \hline\hline
        \multirow{2}{*}{GREY} & GN & 100 & 75 & 88.75 & \textbf{100} & 88.75 & \textbf{99.58} \\ \cline{2-8}
        ~ & GS & 100 & 30 & 95 & 92.08 & 94.58 & 90 \\ \hline\hline
        \multirow{2}{*}{SPRD} & GN & 99.38 & 57.5 & 89.17 & \textbf{99.17} & 89.38 & \textbf{98.75} \\ \cline{2-8}
        ~ & GS & 100 & 21.25 & 92.29 & 85.83 & 90.83 & 83.33 \\ \hline\hline
        \multirow{2}{*}{MASK} & GN & 99.38 & 42.5 & 92.29 & 86.25 & \textbf{90.42} & 89.17 \\ \cline{2-8}
        ~ & GS & 100 & 28.75 & \textbf{92.92} & 74.58 & 90.21 & 76.25 \\ \hline
    \end{tabular}
     \caption{\footnotesize \textbf{Average male and female gender prediction accuracies  on test sets of 480 images of all image types, for fine-tuning on images from the RGB$_{0.3}$ set. Max values for a given combination of image type and learning setting are in bold.}}
    \label{tab:ft_rgb03}
    \end{center}
\end{table}

Next, in Table~\ref{tab:ft_rgb03}, we look at the results for fine-tuning with images from the RGB$_{0.3}$ set. Here we see that for a majority of the combinations females report higher accuracies for both the one-shot and two-shot learning, and males report higher accuracies only in the RGB$_{0.5}$ set for two-shot learning and in the masked set for both one-shot and two-shot. We also note that the highest accuracies reported by females are for those from the Global North. These observations hold true for both one-shot and two-shot learning. Finally we see that the accuracy for females always increases with the few-shot learning (with a higher value in one-shot than in two-shot) with the maximum increase being 70.42\% for females in the Global North in the RGB$_{0.5}$ set.

\begin{table}[!t]
    \tiny
    \begin{center}
    \begin{tabular}{|c|c||c|c||c|c||c|c||}
    \hline
    \multirow{2}{*}{\textbf{Type}} & \multirow{2}{*}{\textbf{Region}} & \multicolumn{2}{|c||}{\textbf{Pre-trained}} & \multicolumn{2}{|c||}{\textbf{1-shot}} & \multicolumn{2}{|c||}{\textbf{2-shot}} \\
     \cline{3-8}
     & & M & F & M & F & M & F \\ \hline \hline
        \multirow{2}{*}{ORIG} & GN & 100 & 70 & 96.25 & 94.58 & 91.04 & \textbf{97.08} \\ \cline{2-8}
        ~ & GS & 100 & 30 & \textbf{97.92} & 65.42 & 94.17 & 80.42 \\ \hline\hline
        \multirow{2}{*}{RGB$_{0.3}$} & GN & 100 & 51.25 & 91.04 & 93.75 & 84.58 & \textbf{97.5} \\ \cline{2-8}
        ~ & GS & 100 & 16.25 & \textbf{96.88} & 64.17 & 91.46 & 85.83 \\ \hline\hline
        \multirow{2}{*}{RGB$_{0.5}$} & GN & 100 & 16.25 & 88.12 & \textbf{90.83} & 83.75 & \textbf{97.08} \\ \cline{2-8}
        ~ & GS & 99.38 & 2.5 & 88.96 & 64.17 & 88.96 & 77.92 \\ \hline\hline
        \multirow{2}{*}{GREY} & GN & 100 & 75 & 96.46 & 92.5 & 90.21 & 92.08 \\ \cline{2-8}
        ~ & GS & 100 & 30 & \textbf{98.54} & 70 & \textbf{94.79} & 67.92 \\ \hline\hline
        \multirow{2}{*}{SPRD} & GN & 99.38 & 57.5 & 95 & 91.67 & 90 & \textbf{95.42} \\ \cline{2-8}
        ~ & GS & 100 & 21.25 & \textbf{96.67} & 60.42 & 93.12 & 76.67 \\ \hline\hline
        \multirow{2}{*}{MASK} & GN & 99.38 & 42.5 & 95.62 & 81.25 & 95.42 & 84.58 \\ \cline{2-8}
        ~ & GS & 100 & 28.75 & \textbf{98.54} & 61.67 & \textbf{96.25} & 68.33 \\ \hline
    \end{tabular}
     \caption{\footnotesize \textbf{Average male and female gender prediction accuracies  on test sets of 480 images of all image types, for fine-tuning on images from the RGB$_{0.5}$ set. Max values for a given combination of image type and learning setting are in bold.}}
    \label{tab:ft_rgb05}
    \end{center}
\end{table}

Similar to the above, in Table~\ref{tab:ft_rgb05}, we look at the results for fine-tuning with images from the RGB$_{0.5}$ set. Here, only the strength of the RGB filter increases but the trend for the improvement in accuracy changes considerably. For one-shot learning, males from Global South report the highest accuracies, except for RGB$_{0.5}$ set. For two-shot learning, females from Global North still report the highest accuracies for a majority of the combinations.

\begin{table}[!t]
    \tiny
    \begin{center}
    \begin{tabular}{|c|c||c|c||c|c||c|c||}
    \hline
    \multirow{2}{*}{\textbf{Type}} & \multirow{2}{*}{\textbf{Region}} & \multicolumn{2}{|c||}{\textbf{Pre-trained}} & \multicolumn{2}{|c||}{\textbf{1-shot}} & \multicolumn{2}{|c||}{\textbf{2-shot}} \\
     \cline{3-8}
     & & M & F & M & F & M & F \\ \hline \hline
        \multirow{2}{*}{ORIG} & GN & 100 & 70 & 86.67 & \textbf{99.58} & 92.71 & \textbf{99.58} \\ \cline{2-8}
        ~ & GS & 100 & 30 & 93.33 & 88.33 & 94.38 & 87.5 \\ \hline\hline
        \multirow{2}{*}{RGB$_{0.3}$} & GN & 100 & 51.25 & 93.13 & 92.08 & 96.67 & 89.58 \\ \cline{2-8}
        ~ & GS & 100 & 16.25 & \textbf{96.46} & 62.5 & \textbf{97.92} & 59.17 \\ \hline\hline
        \multirow{2}{*}{RGB$_{0.5}$} & GN & 100 & 16.25 & 96.04 & 64.58 & \textbf{99.17} & 53.75 \\ \cline{2-8}
        ~ & GS & 99.38 & 2.5 & \textbf{98.54} & 30.42 & \textbf{99.17} & 24.58 \\ \hline\hline
        \multirow{2}{*}{GREY} & GN & 100 & 75 & 86.46 & \textbf{100} & 91.25 & \textbf{99.58} \\ \cline{2-8}
        ~ & GS & 100 & 30 & 94.17 & 91.67 & 94.79 & 90.83 \\ \hline\hline
        \multirow{2}{*}{SPRD} & GN & 99.38 & 57.5 & 87.92 & \textbf{98.33} & 92.29 & \textbf{97.5} \\ \cline{2-8}
        ~ & GS & 100 & 21.25 & 92.29 & 83.75 & 93.33 & 82.92 \\ \hline\hline
        \multirow{2}{*}{MASK} & GN & 99.38 & 42.5 & 90 & 87.92 & \textbf{93.96} & 86.25 \\ \cline{2-8}
        ~ & GS & 100 & 28.75 & \textbf{91.88} & 75.83 & 93.12 & 74.58 \\ \hline
    \end{tabular}
     \caption{\footnotesize \textbf{Average male and female gender prediction accuracies  on test sets of 480 images of all image types, for fine-tuning on images from the GREY set. Max values for a given combination of image type and learning setting are in bold.}}
    \label{tab:ft_grey}
    \end{center}
\end{table}

In Table~\ref{tab:ft_grey}, we look at the results for fine-tuning with images from the greyscale set. Recall that the greyscale filter removes the skintone from the images. Similar to Table~\ref{tab:ft_orig}, the highest accuracies reported by females are for those from the Global North on the ORIG, GREY and SPRD sets, whereas the highest accuracies reported by males are on the two RGB sets and the MASK set. In fact we also see that the maximum increase in accuracy for females is 63.33\% for females in the Global South in the SPRD set for one-shot learning. As with all previous experiments, the accuracy for females is higher in the one-shot scenario and for males in the two-shot scenario.

\begin{table}[!t]
    \tiny
    \begin{center}
    \begin{tabular}{|c|c||c|c||c|c||c|c||}
    \hline
    \multirow{2}{*}{\textbf{Type}} & \multirow{2}{*}{\textbf{Region}} & \multicolumn{2}{|c||}{\textbf{Pre-trained}} & \multicolumn{2}{|c||}{\textbf{1-shot}} & \multicolumn{2}{|c||}{\textbf{2-shot}} \\
     \cline{3-8}
     & & M & F & M & F & M & F \\ \hline \hline
        \multirow{2}{*}{ORIG} & GN & 100 & 70 & 90.62 & \textbf{100} & 89.58 & \textbf{98.75} \\ \cline{2-8}
        ~ & GS & 100 & 30 & 95.83 & 81.25 & 95.21 & 83.75 \\ \hline\hline
        \multirow{2}{*}{RGB$_{0.3}$} & GN & 100 & 51.25 & 92.29 & 92.08 & 93.12 & 90.42 \\ \cline{2-8}
        ~ & GS & 100 & 16.25 & \textbf{97.08} & 60.83 & \textbf{95.62} & 60.83 \\ \hline\hline
        \multirow{2}{*}{RGB$_{0.5}$} & GN & 100 & 16.25 & 96.25 & 65 & 95.83 & 67.92 \\ \cline{2-8}
        ~ & GS & 99.38 & 2.5 & \textbf{97.5} & 31.67 & \textbf{97.71} & 30.83 \\ \hline\hline
        \multirow{2}{*}{GREY} & GN & 100 & 75 & 89.38 & \textbf{100} & 88.12 & \textbf{98.75} \\ \cline{2-8}
        ~ & GS & 100 & 30 & 97.08 & 85.83 & 95.21 & 86.67 \\ \hline\hline
        \multirow{2}{*}{SPRD} & GN & 99.38 & 57.5 & 88.33 & \textbf{97.5} & 86.88 & \textbf{97.5} \\ \cline{2-8}
        ~ & GS & 100 & 21.25 & 92.92 & 82.08 & 91.87 & 80.42 \\ \hline\hline
        \multirow{2}{*}{MASK} & GN & 99.38 & 42.5 & 92.08 & 86.25 & 92.92 & 86.25 \\ \cline{2-8}
        ~ & GS & 100 & 28.75 & \textbf{94.79} & 70.42 & \textbf{93.96} & 73.75 \\ \hline
    \end{tabular}
     \caption{\footnotesize \textbf{Average male and female gender prediction accuracies  on test sets of 480 images of all image types, for fine-tuning on images from the SPRD set. Max values for a given combination of image type and learning setting are in bold.}}
    \label{tab:ft_sprd}
    \end{center}
\end{table}

Table~\ref{tab:ft_sprd} presents the results for fine-tuning with images from the set which has images with the spread adversarial variant. The trends are exactly same as the fine-tuning setting with ORIG images (Table~\ref{tab:ft_orig}) and GREY images (Table~\ref{tab:ft_grey}), both in terms of maximum accuracy as well as the change in accuracy for the two gender groups. 

\begin{table}[!t]
    \tiny
    \begin{center}
    \begin{tabular}{|c|c||c|c||c|c||c|c||}
    \hline
    \multirow{2}{*}{\textbf{Type}} & \multirow{2}{*}{\textbf{Region}} & \multicolumn{2}{|c||}{\textbf{Pre-trained}} & \multicolumn{2}{|c||}{\textbf{1-shot}} & \multicolumn{2}{|c||}{\textbf{2-shot}} \\
     \cline{3-8}
     & & M & F & M & F & M & F \\ \hline \hline
        \multirow{2}{*}{ORIG} & GN & 100 & 70 & 97.5 & 92.08 & 96.46 & 90.83 \\ \cline{2-8}
        ~ & GS & 100 & 30 & \textbf{98.75} & 63.33 & \textbf{98.75} & 65.83 \\ \hline\hline
        \multirow{2}{*}{RGB$_{0.3}$} & GN & 100 & 51.25 & \textbf{99.79} & 81.25 & 97.71 & 80 \\ \cline{2-8}
        ~ & GS & 100 & 16.25 & 98.96 & 38.33 & \textbf{98.96} & 42.5 \\ \hline\hline
        \multirow{2}{*}{RGB$_{0.5}$} & GN & 100 & 16.25 & 98.96 & 51.25 & \textbf{99.38} & 53.33 \\ \cline{2-8}
        ~ & GS & 99.38 & 2.5 & \textbf{99.38} & 19.58 & \textbf{99.38} & 23.75 \\ \hline\hline
        \multirow{2}{*}{GREY} & GN & 100 & 75 & 97.29 & 92.08 & 96.67 & 90.83 \\ \cline{2-8}
        ~ & GS & 100 & 30 & \textbf{99.17} & 66.67 & \textbf{98.54} & 67.5 \\ \hline\hline
        \multirow{2}{*}{SPRD} & GN & 99.38 & 57.5 & 97.92 & 85 & 96.67 & 83.33 \\ \cline{2-8}
        ~ & GS & 100 & 21.25 & \textbf{98.12} & 53.33 & \textbf{98.12} & 54.58 \\ \hline\hline
        \multirow{2}{*}{MASK} & GN & 99.38 & 42.5 & 94.79 & 84.17 & 92.29 & 85.83 \\ \cline{2-8}
        ~ & GS & 100 & 28.75 & \textbf{96.04} & 71.67 & \textbf{92.71} & 77.5 \\ \hline
    \end{tabular}
     \caption{\footnotesize \textbf{Average male and female gender prediction accuracies  on test sets of 480 images of all image types, for fine-tuning on images from the MASK set. Max values for a given combination of image type and learning setting are in bold.}}
    \label{tab:ft_mask}
    \end{center}
\end{table}

In Table~\ref{tab:ft_mask} we see the results for fine-tuning with masked images. Interestingly, here we see that the maximum accuracy is always reported for the male images, independent of the learning setting or the image type being tested. Thus, we see that even though there is an improvement in accuracy for females, the highest accuracy is still achieved for male photos. Thus, having an occlusion-based adversary in the fine-tuning setting does not give as much benefit as other adversarial settings. Another difference we note compared to all other adversarial variants is that for all other fine-tuning settings observed till now, if the fine-tuning and test dataset are the same, then females always report the highest accuracies. The same does not hold for the masked variant.

\begin{table}[!t]
    \tiny
    \begin{center}
    \begin{tabular}{|c|c||c|c||c|c||c|c||}
    \hline
    \multirow{2}{*}{\textbf{Type}} & \multirow{2}{*}{\textbf{Region}} & \multicolumn{2}{|c||}{\textbf{Pre-trained}} & \multicolumn{2}{|c||}{\textbf{1-shot}} & \multicolumn{2}{|c||}{\textbf{2-shot}} \\
     \cline{3-8}
     & & M & F & M & F & M & F \\ \hline \hline
        \multirow{2}{*}{ORIG} & GN & 100 & 70 & 87.5 & \textbf{100} & 91.25 & \textbf{100} \\ \cline{2-8}
        ~ & GS & 100 & 30 & 93.33 & 85 & 95.42 & 84.58 \\ \hline\hline
        \multirow{2}{*}{RGB$_{0.3}$} & GN & 100 & 51.25 & 91.04 & 95 & 93.33 & 92.92 \\ \cline{2-8}
        ~ & GS & 100 & 16.25 & \textbf{94.58} & 72.08 & \textbf{96.88} & 64.17 \\ \hline\hline
        \multirow{2}{*}{RGB$_{0.5}$} & GN & 100 & 16.25 & 93.75 & 73.75 & 95.83 & 70 \\ \cline{2-8}
        ~ & GS & 99.38 & 2.5 & \textbf{96.04} & 37.08 & \textbf{97.29} & 36.67 \\ \hline\hline
        \multirow{2}{*}{GREY} & GN & 100 & 75 & 86.46 & \textbf{100} & 91.25 & \textbf{100} \\ \cline{2-8}
        ~ & GS & 100 & 30 & 94.17 & 91.25 & 96.04 & 87.5 \\ \hline\hline
        \multirow{2}{*}{SPRD} & GN & 99.38 & 57.5 & 87.71 & \textbf{100} & 92.08 & \textbf{97.5} \\ \cline{2-8}
        ~ & GS & 100 & 21.25 & 90.21 & 85.83 & 95 & 76.25 \\ \hline\hline
        \multirow{2}{*}{MASK} & GN & 99.38 & 42.5 & 90 & 88.33 & 93.75 & 84.17 \\ \cline{2-8}
        ~ & GS & 100 & 28.75 & \textbf{91.25} & 75 & \textbf{95.62} & 67.92 \\ \hline
    \end{tabular}
     \caption{\footnotesize \textbf{Average male and female gender prediction accuracies  on test sets of 480 images of all image types, for a fine-tuning setting where 50\% images are chosen from the ORIG set and the other 50\% images are chosen equally from RGB$_{0.3}$, GREY, SPRD and MASK set. Max values for a given combination of image type and learning setting are in bold.}}
    \label{tab:ft_all_adv}
    \end{center}
\end{table}

Finally, we look at the results for the mixed fine-tuning setup as explained previously, in Table~\ref{tab:ft_all_adv}. Again, the trends are same as Table~\ref{tab:ft_orig},~\ref{tab:ft_grey} and~\ref{tab:ft_sprd}, but here we see that a majority of the instances where females report the highest accuracy (all from the Global North), the value itself is 100\%, thereby showing the benefit of fine-tuning using a mixture of images from all variants. In fact the maximum improvement in accuracy for females is 65\%, which is the highest for all variants.

\noindent \textbf{Results for contrastive learning}: One may recall from the main draft, for every anchor point (a data point in a given class) $x_a$, the positive example $x_{a}^{+}$ is the RGB$_{0.3}$ variant of the same image. To choose the negative example $x_{a}^{-}$, we employ two techniques -- (i) $x_{a}^{-}$ is chosen as an image of a person of the opposite gender from the ORIG set with a probability of 1 and, (ii) $x_{a}^{-}$ is chosen as an image of a person of the opposite gender from the ORIG set with a probability of 0.85, and an image from another person of the same gender from the ORIG set with a probability of 0.15. We show the experimental results for both setups in Table~\ref{tab:contrastive_supp}. 

We see that for the (1:0) setting, there is a consistent increase in accuracy for females and a consistent drop in accuracy for males. Interestingly, even with the shift in accuracies, the maximum accuracy for four out of the six image sets are reported by males from the Global South. Females from Global North report the max accuracy in the two RGB sets, thereby showing the benefit of choosing adversarial data as positive examples for the contrastive learning setup. In fact the maximum gain in accuracy for female prediction is observed for the RGB$_{0.5}$ set with a gain of 73.75\%. The drop in accuracy for males is at most 21.25\%, also for the RGB$_{0.5}$ set.

\begin{table}[!ht]
    \tiny
    \begin{center}
    \begin{tabular}{|c|c||c|c||c|c||c|c||}
    \hline
    \multirow{2}{*}{\textbf{Type}} & \multirow{2}{*}{\textbf{Region}} & \multicolumn{2}{|c||}{\textbf{Pre-trained}} & \multicolumn{2}{|c||}{\textbf{1:0}} & \multicolumn{2}{|c||}{\textbf{0.85:0.15}} \\
     \cline{3-8}
     & & M & F & M & F & M & F \\ \hline \hline
        \multirow{2}{*}{ORIG} & GN & 100 & 70 & 82.5 & 88.75 & 79.38 & 90 \\ \cline{2-8}
        ~ & GS & 100 & 30 & \textbf{90} & 80 & 88.12 & \textbf{92.5} \\ \hline\hline
        \multirow{2}{*}{RGB$_{0.3}$} & GN & 100 & 51.25 & 81.88 & \textbf{88.75} & 80.62 & 83.75 \\ \cline{2-8}
        ~ & GS & 100 & 16.25 & 87.5 & 76.25 & \textbf{90} & 77.5 \\ \hline\hline
        \multirow{2}{*}{RGB$_{0.5}$} & GN & 100 & 16.25 & 78.75 & \textbf{87.5} & 81.88 & 71.25 \\ \cline{2-8}
        ~ & GS & 99.38 & 2.5 & 86.88 & 76.25 & \textbf{91.25} & 71.25 \\ \hline\hline
        \multirow{2}{*}{GREY} & GN & 100 & 75 & 81.25 & 77.5 & 85 & 63.75 \\ \cline{2-8}
        ~ & GS & 100 & 30 & \textbf{81.88} & 67.5 & \textbf{86.25} & 66.25 \\ \hline\hline
        \multirow{2}{*}{SPRD} & GN & 99.38 & 57.5 & 81.25 & 86.25 & 75 & 87.5 \\ \cline{2-8}
        ~ & GS & 100 & 21.25 & \textbf{88.12} & 80 & 89.38 & \textbf{92.5} \\ \hline\hline
        \multirow{2}{*}{MASK} & GN & 99.38 & 42.5 & 93.12 & 52.5 & \textbf{95} & 32.5 \\ \cline{2-8}
        ~ & GS & 100 & 28.75 & \textbf{93.75} & 47.5 & \textbf{95} & 56.25 \\ \hline
    \end{tabular}
     \caption{\footnotesize \textbf{Average male and female gender prediction accuracies  on test sets of 480 images of all image types, for contrastive learning. The negative example, $x_{a}^{-}$ can be chosen from the opposite gender with a probability of 1 (middle set of columns) or, with a probability of 0.85 from the opposite gender and 0.15 from the same gender, all from the ORIG set. Max values for a given combination of image type and learning setting are in bold.}}
    \label{tab:contrastive_supp}
    \end{center}
\end{table}

To further investigate a contrastive learning setup that better reflects the gender distribution of the \textsc{FARFace} dataset (85.98\% males and 14.02\% females), we experiment with a setting where the negative example, $x_{a}^{-}$ is chosen with a probability of 0.85 from the opposite gender and 0.15 from the same gender, from the ORIG set. In Table~\ref{tab:contrastive_supp} (last column), we see that while there is a better performance for males from the Global South for almost all image sets, as compared to the previous setting, the improvement in accuracy for females is extremely low, with it reducing in some combinations like GREY and MASK for females from the Global North. Thus, we do not see an absolute monotonic improvement in accuracy for females in this scenario as compared to when all negative examples were chosen from the opposite gender. In future work we plan to use other adversarial sets for the contrastive learning as well.

\end{document}